%% file: neurips_2025.tex
\documentclass{article}

    \usepackage[preprint]{neurips_2025}

\usepackage[table]{xcolor}
\usepackage[utf8]{inputenc} 
\usepackage[T1]{fontenc}    
\usepackage{hyperref}       
\usepackage{url}            
\usepackage{booktabs}       
\usepackage{amsfonts}       
\usepackage{nicefrac}       
\usepackage{microtype}      
\usepackage{times}
\usepackage{latexsym}
\usepackage{inconsolata}
\usepackage{amsmath}
\usepackage{amssymb}
\usepackage{algorithm}
\usepackage{algpseudocode}
\usepackage{xspace}
\usepackage{multirow}
\usepackage{graphicx}
\usepackage{tabularx}
\usepackage{subcaption}
\usepackage{wrapfig}
\usepackage{tcolorbox}

\newcommand{\paratitle}[1]{\vspace{1.5ex}\noindent\textbf{#1}}
\newcommand{\ie}{\emph{i.e.,}\xspace}

\newcommand{\eg}{\emph{e.g.,}\xspace}

\newcommand{\ignore}[1]{}

\newcommand{\OURS}{{\textbf{CAST}}\xspace}

\algnewcommand{\Inputs}[1]{%
  \State \textbf{Inputs:}
  \Statex \hspace*{\algorithmicindent}\parbox[t]{0.98\linewidth}{\raggedright #1}
}
\algnewcommand{\Initialize}[1]{%
  \State \textbf{Initialize:}
  \Statex \hspace*{\algorithmicindent}\parbox[t]{0.98\linewidth}{\raggedright #1}
}

\title{Enhancing Cross-task Transfer of Large Language Models via Activation Steering}

\author{
    \textbf{
        Xinyu Tang\textsuperscript{\rm{1}\thanks{\ \ Equal contribution.}},
        Zhihao Lv\textsuperscript{\rm{1}\footnotemark[1]},
        Xiaoxue Cheng\textsuperscript{\rm{1}\ },
        Junyi Li\textsuperscript{\rm{2}\ },
    } \\
    \textbf{
        Wayne Xin Zhao\textsuperscript{\rm{1}\thanks{\ \ Corresponding author.}},
        Zujie Wen\textsuperscript{\rm{3}\ },
        Zhiqiang Zhang\textsuperscript{\rm{3}\ },
        Jun Zhou\textsuperscript{\rm{3}\ }
    } \\
    \textsuperscript{1}Gaoling School of Artificial Intelligence, Renmin University of China \\
    \textsuperscript{2}Department of Computer Science, National University of Singapore
    \textsuperscript{3}Ant Group \\
    \texttt{txy20010310@163.com, joyboy15894@gmail.com}
}

\begin{document}

\maketitle

\input{sections/abstract}

\input{sections/introduction}

\input{sections/related_work}

\input{sections/empirical_study}

\input{sections/method}

\input{sections/experiments}

\input{sections/conclusion}

\bibliographystyle{unsrt}
\bibliography{newbib}

\appendix

\newpage

\input{sections/appendix/algorithm}

\input{sections/appendix/experiments}

\input{sections/appendix/datasets}

\input{sections/appendix/limitation}

\input{sections/appendix/case_study}

\end{document}

%% file: sections/abstract.tex
\begin{abstract}

Large language models~(LLMs) have shown impressive abilities in leveraging pretrained knowledge through prompting, but they often struggle with unseen tasks, particularly in data-scarce scenarios.
While cross-task in-context learning offers a direct solution for transferring knowledge across tasks, it still faces critical challenges in terms of robustness, scalability, and efficiency.
In this paper, we investigate \textit{whether cross-task transfer can be achieved via \textbf{latent space steering}} without parameter updates or input expansion.
Through an analysis of activation patterns in the latent space of LLMs, we observe that the enhanced activations induced by in-context examples have consistent patterns across different tasks.
Inspired by these findings, we propose \OURS, a novel \textbf{C}ross-task \textbf{A}ctivation \textbf{S}teering \textbf{T}ransfer framework that enables effective transfer by manipulating the model's internal activation states.
Our approach first selects influential and diverse samples from high-resource tasks, then utilizes their contrastive representation-enhanced activations to adapt LLMs to low-resource tasks.
Extensive experiments across both cross-domain and cross-lingual transfer settings show that our method outperforms competitive baselines and demonstrates superior scalability and lower computational costs.
  
\end{abstract}

%% file: sections/introduction.tex
\section{Introduction}
\label{sec-intro}

Large language models~(LLMs)~\cite{LLM-survey,openai-o1,deepseek-r1} have demonstrated remarkable capabilities to acquire and store knowledge during pretraining, which can be effectively accessed through prompting.
However, as this usage becomes more prevalent, a crucial challenge emerges: these models often struggle with tasks that were not seen during pretraining, especially in data-scarce scenarios~\cite{Inherent-limits}.
One common approach to tackle this issue is transfer learning~\cite{transfer-learning1,transfer-learning2,transfer-learning3}, which uses knowledge from high-resource tasks to adapt to low-resource ones. 
Some studies~\cite{continuous1,continuous2,continuous3,continuous4} focus on fine-tuning soft-prompts for cross-task transfer.
They typically train soft prompts on data-sufficient tasks, and then apply them to data-scarce tasks by adding them to the input during inference.
While effective, these methods still require training and can not generalize well across diverse tasks.
An alternative approach is cross-task in-context learning~(ICL)~\cite{cross-lingual-icl,discrete2,discrete3,cross-domain-ICL}, which leverages labeled examples from high-resource tasks to improve performance on low-resource tasks without parameter updates.

Despite its promise, cross-task ICL still suffers from several limitations:
(1) Its performance is highly sensitive to the choice of demonstrations~\cite{Demo-Selection1,Demo-Selection2,Demo-Selection3}, prompt templates~\cite{Template1,Template2}, and source tasks~\cite{cross-domain-ICL}, which restricts its adaptability~(robustness).
(2) Few demonstrations can be included due to the limited context length of LLMs~(scalability).
(3) Higher computational costs with an increasing number of demonstrations, for the quadratic computational complexity of Transformers to the input length~(efficiency).
To address these challenges, one possible solution is to operate in the continuous space~\cite{continuous-learning1,continuous-learning2}.
This approach does not require adding explicit input tokens, thereby avoiding increasing the input context length.
This leads us to raise a research question: \textit{Can we achieve effective cross-task transfer through \textbf{latent space steering}?}

To answer this question, we conduct an empirical study of activation patterns in both zero-shot and few-shot settings.
Our findings reveal that few-shot prompts yield higher matrix entropy than zero-shot prompts, suggesting that LLMs encode richer and more diverse features from in-context examples.
Additionally, we observe that the activation differences between few-shot and zero-shot prompts are nearly parallel across tasks in the model's latent space.
This phenomenon indicates that the enhanced information induced by in-context examples follows consistent patterns across tasks, which provides strong support for the possibility of cross-task transfer in the latent space.

Building on these insights, we propose a novel \textbf{C}ross-task \textbf{A}ctivation \textbf{S}teering \textbf{T}ransfer framework, namely \OURS.
We first select a representative subset of high-resource examples that balance influence and diversity, aiming to improve the efficiency of feature extraction.
Specifically, we first construct a similarity-based graph of high-resource examples, then iteratively choose a sample with the highest combined score of influence and diversity at each step, and repeat this process to form the subset.
Next, we compute the activation differences between few-shot and zero-shot prompts for these selected samples, which are subsequently injected into the forward pass of low-resource queries via activation steering to guide LLMs toward effective cross-task transfer.
This approach is both efficient and scalable, as it leverages pre-extracted activations from high-resource tasks during inference without parameter updates or input expansion.
To validate the effectiveness of our proposed method, we conduct extensive experiments in both cross-domain and cross-lingual scenarios. 
Experimental results demonstrate that our approach consistently outperforms competitive baselines while maintaining lower computational cost.

Our main contributions can be summarized as follows:

$\bullet$ To the best of our knowledge, we are the first to analyze the zero-shot and few-shot prompts from the perspective of activations, and find that the enhanced representation from in-context examples exhibits consistent patterns across tasks in the latent space of LLMs.

$\bullet$ We propose a novel cross-task transfer learning framework based on activation steering without modifying the parameters or increasing the input length, which effectively mitigates the challenge of data scarcity in real-world scenarios.

$\bullet$ Extensive experiments validate the effectiveness of our proposed method in both cross-domain and cross-lingual scenarios, while also showing strong scalability and low computational cost.

%% file: sections/related_work.tex
\section{Related Work}

\paratitle{Transfer Learning.}
In data-scarce scenarios, transfer learning offers a promising solution to alleviate the lack of labeled data in low-resource tasks by leveraging labeled samples from high-resource tasks.
Existing transfer learning approaches for LLMs can be mainly categorized into two types: continuous and discrete cross-task transfer.
Continuous cross-task transfer methods~\cite{continuous1,continuous2,continuous3} learn a shared continuous soft prompt from source tasks and apply it to the target tasks. 
While effective, they require fine-tuning and generalize poorly across diverse tasks.
In contrast, discrete cross-task transfer methods~\cite{cross-lingual-icl,discrete2,discrete3,cross-domain-ICL} directly incorporate high-resource examples into LLM inputs to solve low-resource tasks without parameter updates. 
However, such an approach faces several limitations: its performance depends heavily on the quality of demonstration, the fixed context window constrains scalability, and the quadratic complexity reduces efficiency with longer inputs.
To overcome these issues, we propose a novel approach to extract activations from high-resource tasks and inject them into low-resource tasks.
This approach eliminates the need for fine-tuning or input expansion, while preserving robustness, scalability, and computational efficiency.

\paratitle{Activation Steering.}
Activation steering is a well-established method that treats internal representations as the fundamental unit, focusing on analyzing and manipulating them within neural networks.
This technique has been successfully applied in various scenarios, including alignment~\cite{re-alignment}, personality modeling~\cite{re-person}, instruction following~\cite{re-instruction}, hallucination mitigation~\cite{re-halu1,re-halu2}, safety enhancement~\cite{re-safety}, and reasoning improvement~\cite{re-cot1,re-cot2}.
In this work, we leverage the activation steering method to transfer knowledge from the data-sufficient tasks to address data-scarce tasks.

%% file: sections/empirical_study.tex
\section{Empirical Study}
\label{sec-empirical-study}

The Hopfieldian view of cognition~\cite{hopfield} suggests that neural computation arises from dynamic transformations in population-level neural activity driven by external stimuli.
This perspective has inspired mechanistic interpretations of artificial neural networks, where activation steering~\cite{Representation-Engineer} has emerged as a powerful method for analyzing the internal states of models.
By treating intermediate activations as the fundamental units of computation, activation steering reveals the high-level concepts and functions encoded within models.
In this section, we analyze zero-shot and few-shot prompting in both in-domain and cross-domain settings from the perspective of activations.

\subsection{In-domain Activation Analysis}

\begin{figure}[t]
    \centering
    \begin{minipage}[c]{0.525\textwidth}
        \centering
        \begin{subfigure}[b]{0.49\textwidth}
            \centering
            \includegraphics[width=\textwidth]{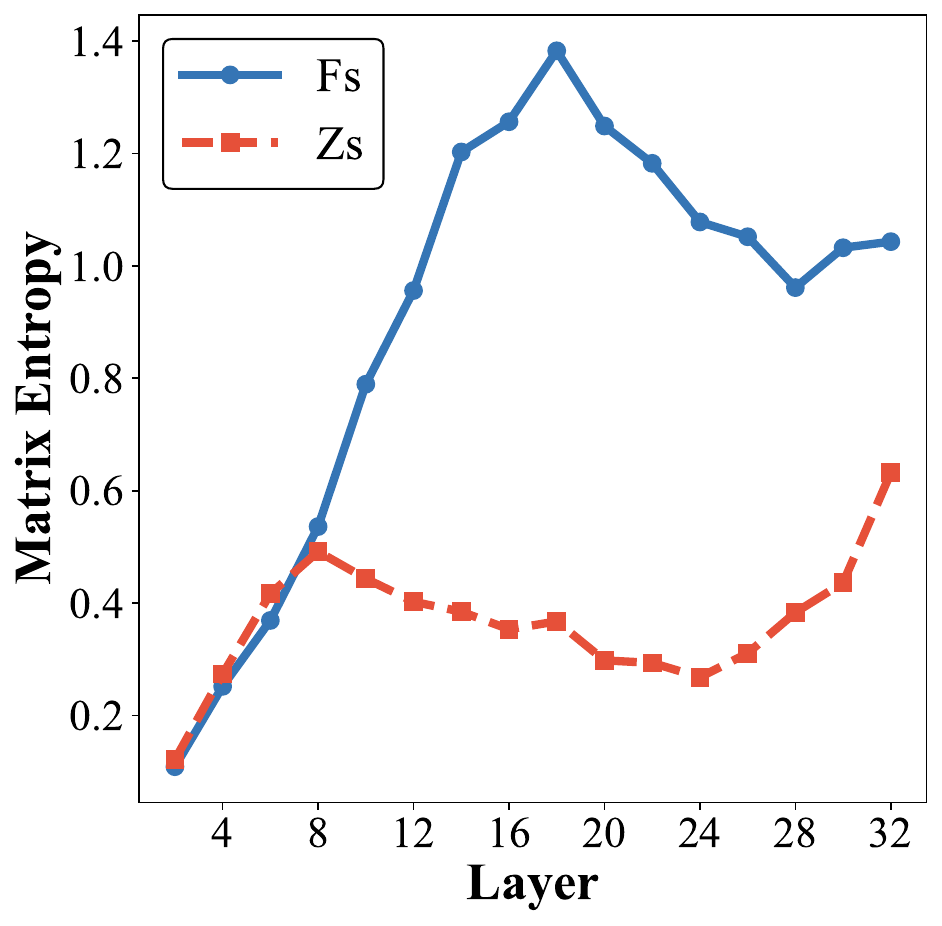}
            \caption{AGnews}
            \label{fig:entropy_agnews}
        \end{subfigure}
        \begin{subfigure}[b]{0.49\textwidth}
            \centering
            \includegraphics[width=\textwidth]{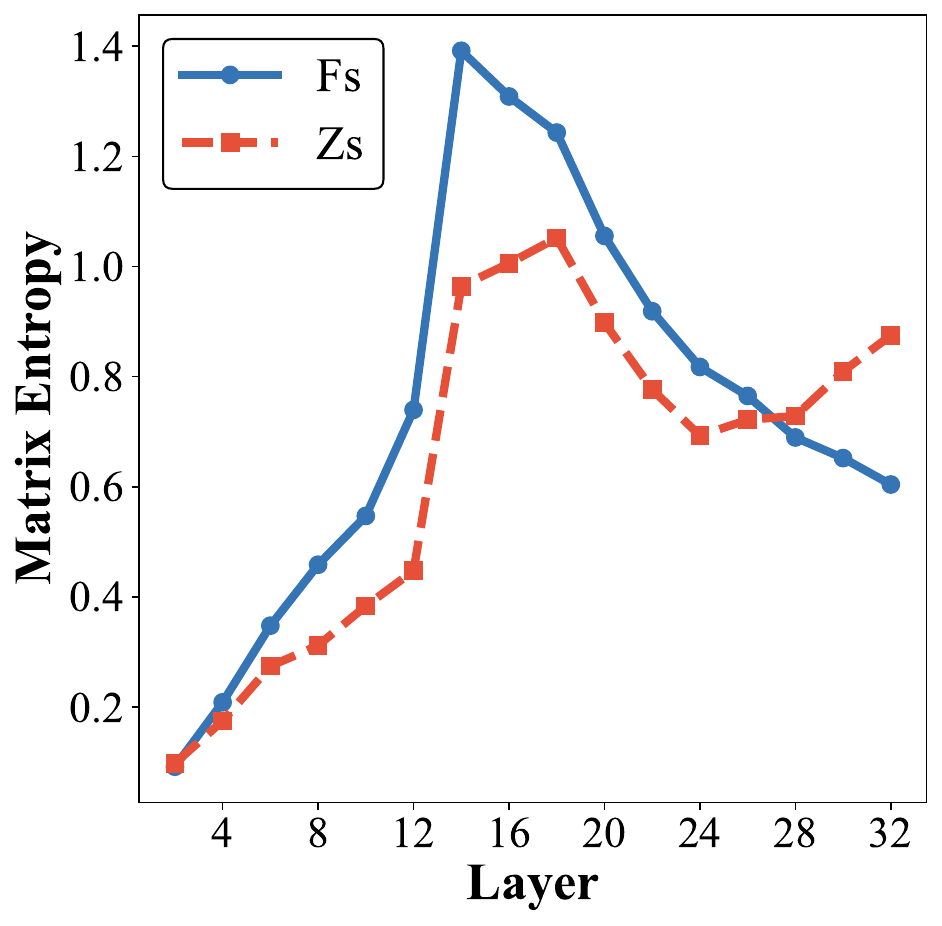}
            \caption{Arc-Easy}
            \label{fig:entropy_arceasy}
        \end{subfigure}
        \caption{Matrix entropy}
        \label{fig:matrix_entropy}
    \end{minipage}
    \hfill
    \begin{minipage}[c]{0.45\textwidth}
        \centering
        \begin{subfigure}[b]{0.49\textwidth}
            \centering
            \includegraphics[width=\textwidth]{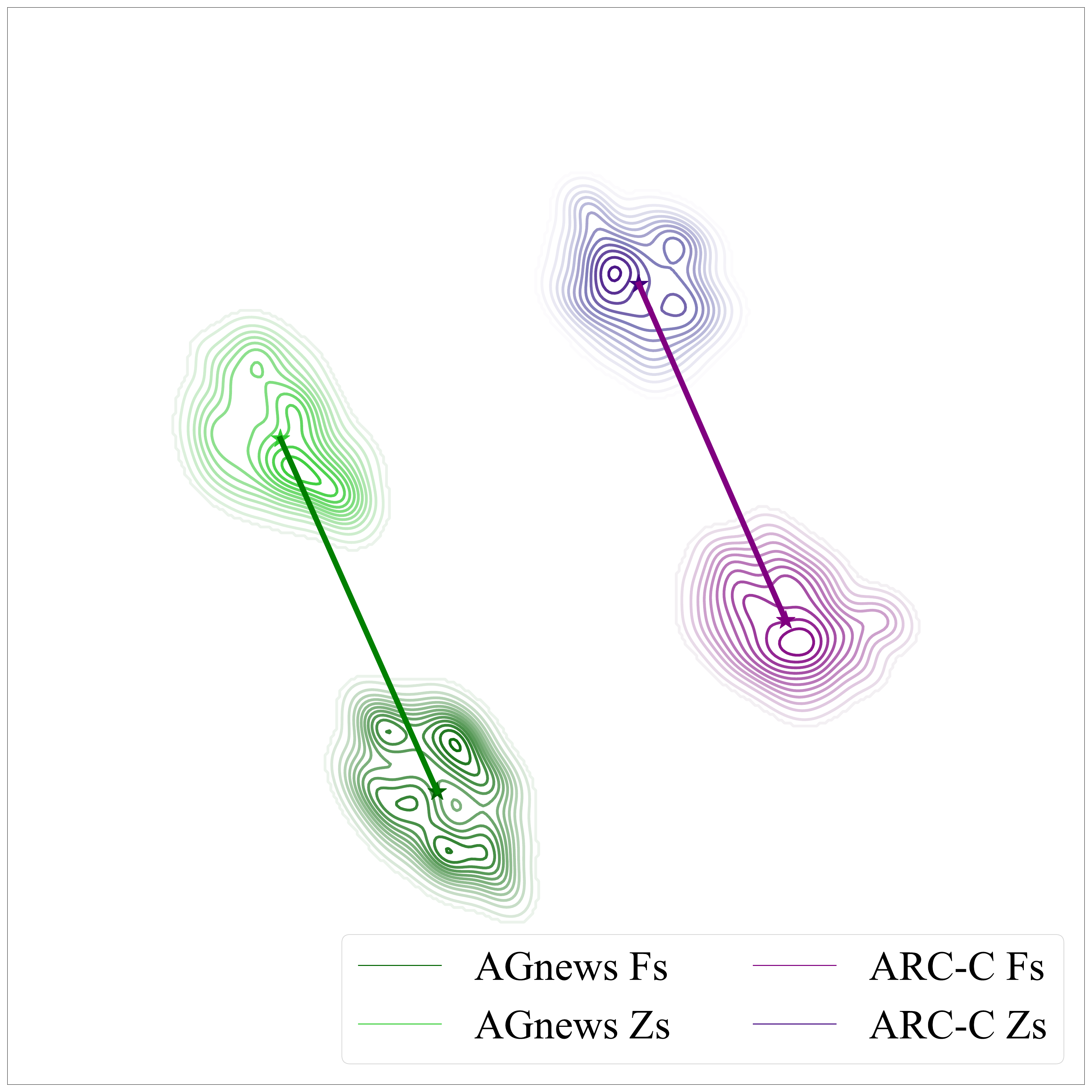}
            \caption{ARC-C \& AGnews}
            \label{fig:a-a}
        \end{subfigure}
        \begin{subfigure}[b]{0.49\textwidth}
            \centering
            \includegraphics[width=\textwidth]{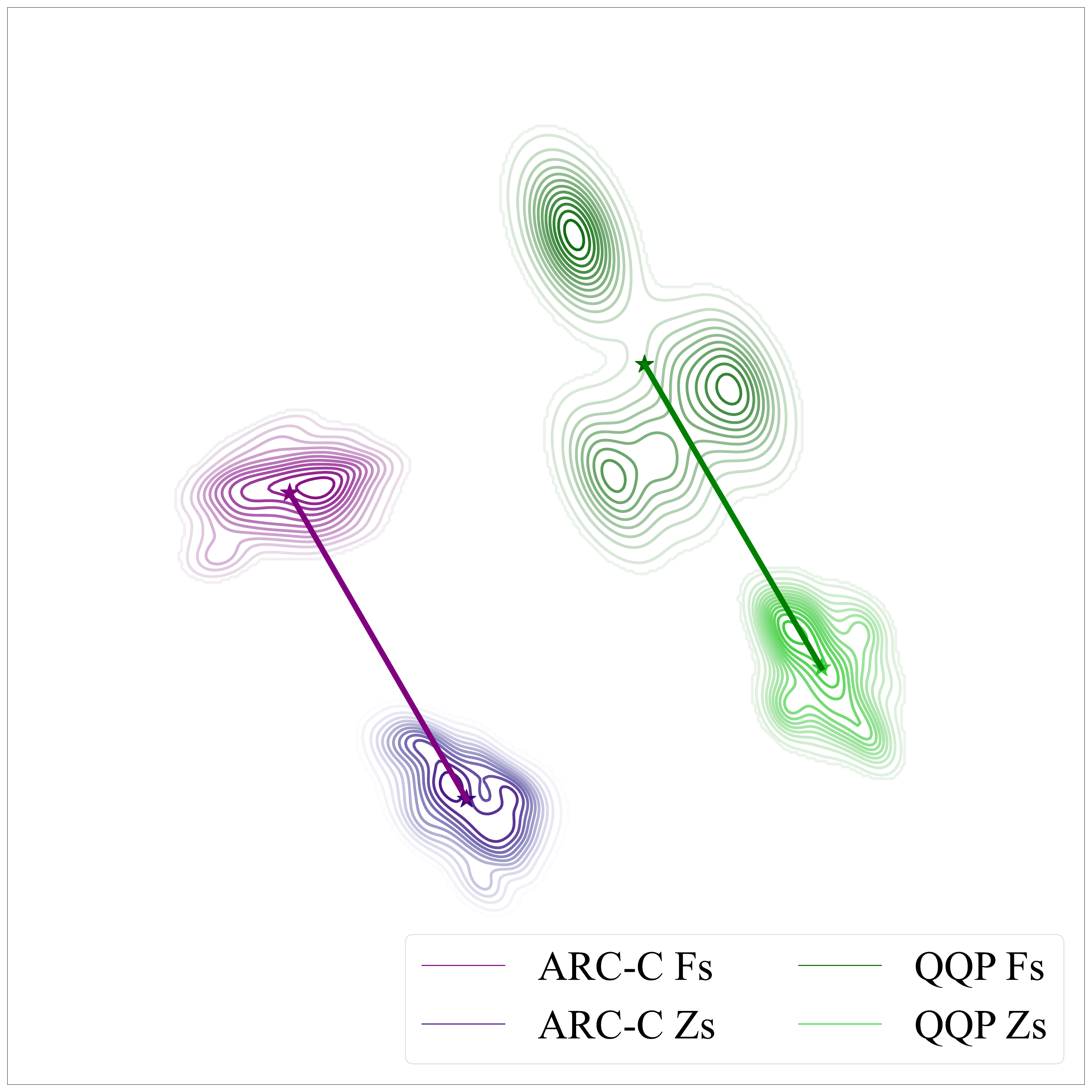}
            \caption{ARC-C \& QQP}
            \label{fig:a-q}
        \end{subfigure}
        \caption{t-SNE cluster across domains ({green} for source task, {purple} for target task)}
        \label{fig:cluster}
    \end{minipage}
\end{figure}

In this section, we analyze zero-shot and few-shot prompting in in-domain scenarios from the perspective of activations.
Specifically, we extract activations from the final token's hidden states, as they aggregate the most contextually salient information in LLMs.
To quantify the structural properties of these activations, we use \textit{matrix-based entropy}~\cite{DiME,Diff-eRank,layer-by-layer} to measure the diversity and redundancy of contextual features within the embedding space.
Higher entropy indicates greater diversity and lower redundancy features, while lower entropy suggests stronger compression effects.

Formally, given a set of $N$ samples embedded in a $D$-dimensional space, represented as $\mathbf{Z} \in \mathbb{R}^{N \times D}$, the matrix-based entropy is comoputed as:
\begin{equation}
    S_{\alpha}(\mathbf{Z}) = \frac{1}{1 - \alpha} \log \left( \sum_{i=1}^{r} \left( \frac{\lambda_i(\mathbf{K})}{\mathrm{tr}(\mathbf{K})} \right)^{\alpha} \right),
\end{equation}
where $\mathbf{K}$ denotes the Gram matrix of $\mathbf{Z}$, $\{\lambda_i(\mathbf{K})\}$ are the nonnegative eigenvalues of $\mathbf{K}$, and $r=\text{rank} (\mathbf{K}) \le min(N, D)$.
Following previous work~\cite{layer-by-layer}, we set $\alpha \to 1$.

As shown in Figure~\ref{fig:matrix_entropy}, our analysis reveals two key findings.
First, few-shot prompting yields higher entropy across almost all layers compared to zero-shot prompting. 
This suggests that LLMs encode richer and more diverse features from in-context examples.
Second, we observe that the entropy of few-shot prompts peaks in intermediate layers, indicating that these layers capture the most semantically rich representations within the models.
This phenomenon aligns with existing findings~\cite{layer-by-layer} that intermediate layers can encode even richer representations than final layers.

\subsection{Cross-domain Activation Analysis}
\label{subsec-cross-domain}

Prior work~\cite{cross-domain-ICL} has demonstrated that incorporating relevant examples from high-resource tasks in the discrete token space can enhance performance on low-resource tasks.
However, due to the limited context window of LLMs, only a small subset of examples can be included in the input, which restricts the full utilization of available data. 
One possible alternative is to store features from high-resource tasks in a continuous representation space to address low-resource tasks, thereby overcoming the input length constraints of the model.
In this part, we investigate whether activations from the high-resource tasks can facilitate cross-task transfer through continuous space manipulation in the latent space of LLMs.

To explore this possibility, we analyze how zero-shot and few-shot prompting across domains are distributed within the model's latent space.
Specifically, we use t-SNE~\cite{t-sne} dimensionality reduction method to map activations onto a 2D plane.
As presented in Figure~\ref{fig:cluster}, activations from these two prompting methods within the same domain form separate clusters, indicating that LLMs generate fundamentally different internal representations under each paradigm.
Furthermore, the difference vectors between these two prompting activations across domains are nearly parallel, suggesting that the enhanced information induced by in-context examples follows a consistent direction across domains.
This observation motivates us to use these contrastive representation-enhanced activations to transfer knowledge from high-resource to low-resource tasks in the latent space.

%% file: sections/method.tex
\section{Method}
\label{sec-method}

\begin{figure*}[t]
    \centering
    \includegraphics[width=\textwidth]{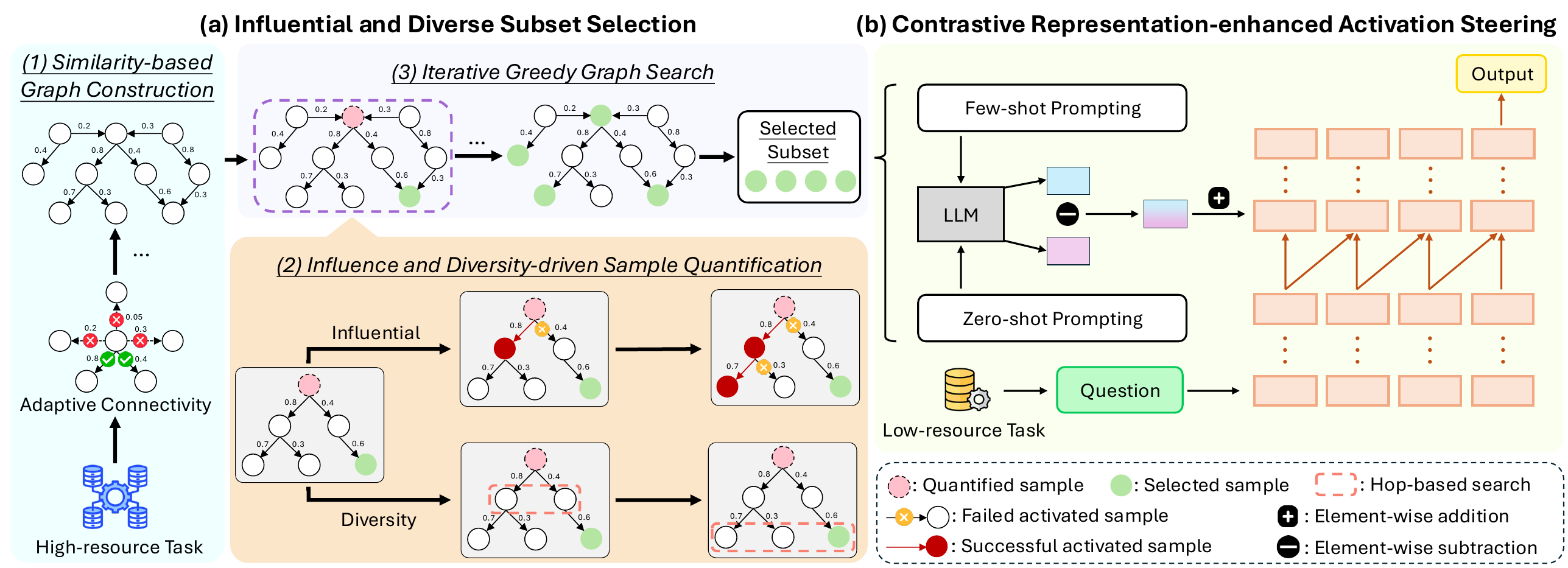}
    \caption{The overview of \OURS. (a) An illustration of three phases in our influential and diverse subset selection strategy. (1) We first construct a similarity-based sample graph of the high-resource task. (2) Next, we quantify each sample's influence and diversity. (3) Then, we use an iterative greedy graph search algorithm to choose the highest-scoring sample at each step, and finally form the subset of the high-resource task. (b) We utilize contrastive representation-enhanced activations of these samples to guide the adaptation of LLMs to the low-resource task within the latent space.}
\label{fig:main}
\end{figure*}

In this section, we introduce \OURS, a novel framework for cross-task transfer via activation steering in LLMs.
We first formalize the problem of cross-task transfer learning, then select influential and diverse samples from high-resource tasks, and finally leverage contrastive representation-enhanced activations from these samples to adapt LLMs to low-resource tasks through activation steering.
The overall framework is illustrated in Figure~\ref{fig:main}.

\subsection{Problem Formulation}

Cross-task transfer learning aims to enable LLMs to leverage knowledge and information from high-resource source tasks to improve performance on low-resource target tasks~\cite {continuous1,cross-domain-ICL}.
Let $D_s$ denote the source task with abundant labeled data, and $D_t$ denote the target task with limited labeled samples.
The process typically involves extracting transferable information $I_s$ from source tasks and applying it to the target task, which can be formalized as:
\begin{align}
    I_s &= f(d_s,x_{s_1},y_{s_1},\cdots ,x_{s_n},y_{s_n}), \\
    \hat{y_t} &= \text{LLM}(I_s,d_t,x_t).
\end{align}
Here, $f$ is the feature extraction function that captures knowledge from the source task, $d_s$ and $d_t$ are the task definitions of the source and target tasks, respectively, $(x_{s_i},y_{s_i})_{i=1}^n \in D_s$ represents the labeled examples from the source task, and $x_t \in D_t$ is an input question from the target task.

\subsection{Influential and Diverse Subset Selection}

Although the source task contains abundant labeled data, using all samples for cross-task transfer is both computationally inefficient and often unnecessary.
To address this, we propose an influential and diverse subset selection strategy that effectively identifies a representative and diverse subset capturing the essential characteristics of the source task.
Our approach consists of three stages: we first build a sample graph based on pairwise similarities to model the relationships among examples, then measure the influence and diversity score of each sample, and employ an iterative greedy search algorithm to select the highest-scoring sample at each step to build the final subset.

\subsubsection{Similarity-based Graph Construction}

Our subset selection strategy begins by modeling the relationships among samples through a directed graph.
To this end, we first encode each sample into a vector using the BGE model~\cite{BGE-model}.
These embeddings are then used to construct a task-specific directed graph $\mathcal{G}=(\mathbf{V},\mathbf{E},\mathbf{P})$, where each vertex $v_i \in \mathbf{V}$ denotes a sample, a directed edge $e(i,j) \in \mathcal{E}$ connects node $v_i$ to its neighbor $v_j$, and edge weight $p(i,j) \in \mathbf{P}$ is determined by the cosine similarity between the corresponding sample embeddings.
To reduce structural redundancy and complexity, we introduce an adaptive connectivity mechanism that dynamically controls the sparsity of the graph.
Specifically, for each node, we dynamically determine the number of edges it connects with its most similar neighbors based on the average similarity to other samples.
This ensures that nodes with higher average similarity establish more connections, while peripheral nodes have fewer edges.
Formally, we define the out-degree of node $v_i$ as $k_i$:
\begin{align}
    s_i &= \frac{1}{|\mathbf{V}|-1}\sum_{j \in \mathbf{V} \cup j \neq i} s(i,j), \\
    k_i &= \lceil \alpha \cdot s_i \cdot (|\mathbf{V}|-1) \rceil,
\end{align}
where $s(i,j)$ denotes the cosine similarity between node $v_i$ and $v_j$, and $\alpha$ is a hyperparameter controlling the sparsity of the graph.

\subsubsection{Influence and Diversity-driven Sample Quantification}

After constructing the sample graph, we quantify each sample from two aspects: (1) its influence in activating other samples within the task, and (2) its contribution to the overall diversity of the previously selected samples.
The corresponding pseudo-code is provided in Algorithm~\ref{alg:quantification}.

The influence score measures how influential a sample is within the task, which is computed by simulating an information diffusion process.
Specifically, we initialize the process by adding the candidate node $v$ into an active set $S_{\text{active}}$.
At each propagation iteration, we randomly select an active node $u \in S_{\text{active}}$ and attempt to activate its neighbors.
Specifically, for each 1-hop neighbor $w \in N_1(u)$, activation succeeds with probability $p(u,w)$, which corresponds to the edge weight between node $u$ and node $w$.
Newly activated nodes that have not been activated before are added to $S_{\text{active}}$.
This iterative process continues until no nodes are in $S_{\text{active}}$.
The influence score $I(v)$ of node $v$ is defined as the total number of nodes activated during the entire diffusion process.
In this way, samples that trigger widespread activation receive higher scores.
To ensure the robustness and reduce variance, we repeat this simulation 20 times and report the average influence score.

The diversity penalty quantifies how much a node would reduce the overall diversity of the current selected subset $S_{\text{selected}}$.
Specifically, we employ a hop-based search method to explore the $i$-hop neighbor $N_i(v)$ of node $v$, and measure the overlap with the already selected nodes $S_{\text{selected}}$.
The diversity penalty $D(v)$ of node $v$ is formulated as:
\begin{align}
    D(v) &= -\sum_{i=1}^k \beta^i\cdot |N_i(v) \cap S_{\text{selected}}|, \\
    f_{\mathcal{G}}(v) &= I(v) + \gamma \cdot D(v).
\end{align}
Here, $\beta$ is a hop-based decay factor that reduces the penalty for overlaps occurring at greater distances from node $v$.
Consequently, nodes with minimal overlap with the existing subset have a smaller diversity penalty.
Finally, we combine the influence score $I(v)$ and diversity penalty $D(v)$ of node $v$ using a balancing hyperparameter $\gamma$, which controls the trade-off between them.

\subsubsection{Iterative Greedy Graph Search}

To construct a subset that effectively captures both task representativeness and sample diversity, we propose an iterative greedy graph search strategy.
The pseudo-code is provided in Algorithm~\ref{alg:selection}.
Our approach operates on the sample graph $\mathcal{G}$ and iteratively selects the candidate sample with the highest influence-diversity score.
In more detail, the selection process begins with an empty set.
At each iteration, we compute the influence-diversity score for all unselected samples using the sample node evaluation function $f_{\mathcal{G}}$.
The sample with the highest score is then added to the subset.
This procedure is repeated until the desired subset size is reached.

Notably, our iterative subset selection process is highly efficient, as we do not need to re-evaluate all nodes from scratch at each iteration.
The influence scores are precomputed and remain unchanged throughout the search process.
Only the diversity penalties of those samples connected to the most recently selected node need to be updated.

\subsection{Contrastive Representation-enhanced Activation Steering}

Inspired by the observation that the difference vectors between zero-shot and few-shot promptings exhibit consistent patterns across tasks in the latent space~(in Section~\ref{subsec-cross-domain}), we propose an activation steering method that leverages these contrastive representations to transfer high-resource information for cross-task transfer from high-resource tasks to low-resource ones.
Our method primarily consists of two components: \textit{activation extraction} and \textit{activation control}.

\paratitle{Activation Extraction.}
The component aims to identify high-level concepts or functional behaviors encoded in LLMs.
In this part, we aim to extract the contrastive representation-enhanced activations from high-resource tasks.
Specifically, for each sample $s_i \in S_n$ from the high-resource task, we construct two types of prompts: a zero-shot prompt $z_i$ which contains only the sample itself, and a few-shot prompt $f_i$ which includes the sample along with similar demonstrations.
To eliminate instance-specific noise and capture general task-level features, we compute the average activation difference across all samples.
Formally, we define the high-resource contrastive representation-enhanced activations $C_L$ as:
\begin{equation}
    C_L = \frac{1}{n} \sum_{i=1}^{n} (A_L(f_i) - A_L(z_i)),
\end{equation}
where $A_L$ denotes the operation of extracting the activation from the final token's hidden state at layer $L$, and $n$ is the number of samples.

\paratitle{Activation Control.}
The component aims to steer model behaviors using the extracted activations.
In this work, we inject the high-resource contrastive representation-enhanced activations $C_L$ into the forward pass on low-resource questions during inference, guiding the model toward effective cross-task transfer.
Specifically, we follow prior work~\cite{re-safety} to inject the activations at the final token position of a specific layer, allowing for a strong influence on the output generation while ensuring minimal disruption to the encoding of earlier tokens.
The modified hidden state is computed as:
\begin{equation}
    \hat{h}_L^{-1} = h_L^{-1} + \lambda \cdot C_L ,
\end{equation}
where $\lambda$ controls the activation injection strength, and $h_L^{-1}$ is the final token's hidden state at layer $L$.

%% file: sections/experiments.tex
\section{Experiments}
\label{sec-experiments}

\input{tables/main_cd}
\input{tables/name_cl}


\subsection{Experimental Setup}

\paratitle{Datasets.}
In this paper, we evaluate our method in both cross-domain and cross-lingual scenarios.
For the cross-domain experiments, we use seven source domains and five target domains.
The source domains include: ARC-Easy~\cite{ARC}, AG-news~\cite{AG-news}, BoolQ ~\cite{BoolQ}, Commonsense-QA~\cite{Commonsense-QA}, MNLI~\cite{MNLI}, QQP~\cite{QQP}, and SST2~\cite{SST2}.
Following previous work~\cite{cross-domain-ICL}, we select ARC-Challenge~\cite{ARC}, Financial-Phrasebank~\cite{Financial-Phrasebank}, MedMCQA ~\cite{MedMCQA}, SciQ~\cite{SciQ}, and Social-i-QA~\cite{Social-i-QA} as target domains.
Among these, ARC-Challenge is a more difficult version of the source domain ARC-Easy, while the remaining four require domain-specific knowledge.
In the cross-lingual settings, we conduct experiments on the MARC~\cite{MARC} dataset, which spans six languages.
Due to computational constraints, we follow previous work~\cite{cross-domain-ICL} and randomly sample 500 examples from each target domain as the test set.
Detailed descriptions of the datasets are provided in Appendix~\ref{app:dataset}.

\paratitle{Baselines.}
We select several training-free approaches for comparison, including zero-shot and several representative few-shot prompting methods.
(1) \underline{\textit{Zero-shot}} prompting method takes only the target domain question as input, without any demonstrations.
(2) \underline{\textit{Few-shot Random}} prompting method randomly selects examples from the source domain and prepends them to the target question as input.
(3) \underline{\textit{Few-shot TopK}}~\cite{Fs-TopK} prompting method retrieves similar examples from the source domain to the target question as context.
(4) \underline{\textit{Few-shot DPP}}~\cite{Fs-DPP} prompting method uses the determinantal point process approach to select diverse demonstrations as input.

\paratitle{Implementation Details.}
Our main experiments are conducted using Llama3.1-8B-Instruct~\cite{Llama}.
Additional Experiments using different LLMs are provided in Appendix~\ref{app-sub:different-llm}.
In our subset selection strategy, we set the size of the selected subset $n$ to 20, the hyperparameter $\beta$ controlling the sparsity of the graph to 20, the hop-based decay factor $\alpha$ to 0.2, and the trade-off parameter $\gamma$ between influence and diversity to 0.2.
For activation steering, we inject extracted activations at the hidden state of the final token position. 
The injection layer is determined based on performance on the validation set, and the injection strength $\lambda$ is set to 1.
We use accuracy as our evaluation metric for the experiments.
All the experiments are computed on 8 A800 GPUs.

\subsection{Experimental Results}

\paratitle{Cross-domain Scenarios.}
Table~\ref{tab:main-cd} shows the results for the cross-domain transfer setting.
Among the demonstration selection methods, TopK and DPP demonstration selection method achieve slight improvements over random selection by selecting examples that are more relevant and diverse.
However, we observe that the effectiveness of the cross-task few-shot prompting is highly dependent on the similarity between the source and target domains.
When the source and target domains are closely related (\eg ARC-Easy to ARC-Challenge), incorporating examples from the source domain can improve performance. 
In contrast, for dissimilar domain pairs (\eg ARC-Easy to MedMCQA), such examples can introduce noise and lead to performance degradation.
Furthermore, our proposed method consistently outperforms all baselines across all domain pairs, even when the source and target domains are not similar.
This superior performance stems from the fact that the enhanced information induced by in-context examples from different domains shares consistent directions in the latent space.
By injecting representation-enhanced activations from high-resource domains into the latent space during the forward pass of the low-resource domains, our approach enables effective cross-domain transfer without introducing input-level noise.

\paratitle{Cross-lingual Scenarios.}
Table~\ref{tab:main-cl} presents the results for the cross-lingual transfer setting.
We find that the performance of cross-lingual few-shot prompting varies significantly depending on the linguistic similarity between the source and target languages.
For closely related language pairs (\eg French $\to$ Spanish), incorporating cross-lingual examples generally improves performance.
However, for distant language pairs (\eg English $\to$ Japanese), such demonstrations may introduce noise and result in negative transfer.
Our method consistently outperforms all baselines and demonstrates strong generalization across a wide range of languages. 
For example, it achieves an average accuracy of 92.28\% for German (compared to 87.76\% for DPP) and 89.04\% for English (compared to 64.48\% for DPP), highlighting its robust cross-lingual transfer capability.
This improvement can be attributed to the extraction of informative activations from high-resource languages, which are injected into the forward pass of low-resource languages within the latent space.

\subsection{Detailed Analysis}

In this part, we conduct detailed analysis of the generalization, scalability, efficiency and effectiveness of \OURS.
Unless otherwise stated, we use MNLI and MedMCQA as source and target tasks.

\subsubsection{Experiments on Different-scale LLMs}

\begin{figure}[t]
    \centering
    \begin{subfigure}[b]{0.32\columnwidth}
        \centering
        \includegraphics[width=\columnwidth]{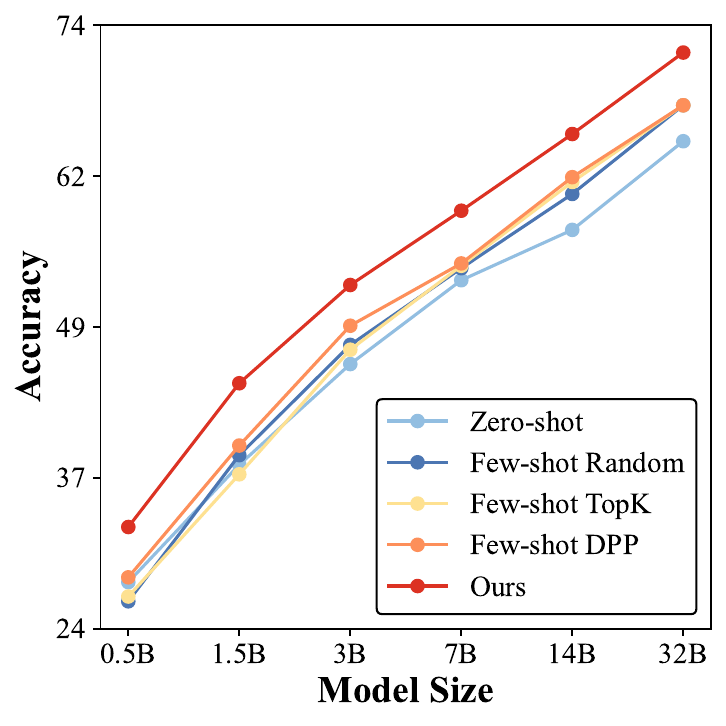}
        \caption{Different Model Sizes}
        \label{fig:different_llms}
    \end{subfigure}
    \begin{subfigure}[b]{0.32\columnwidth}
        \centering
        \includegraphics[width=\columnwidth]{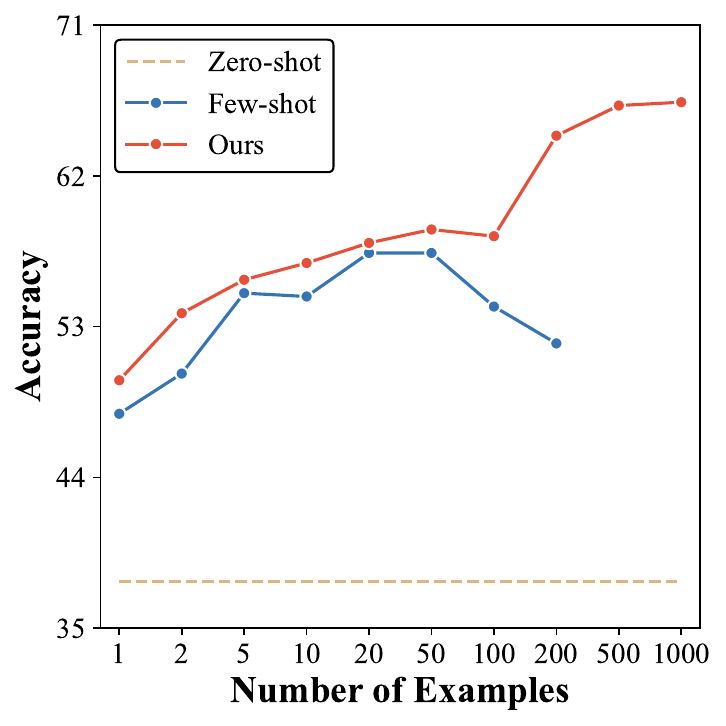}
        \caption{Scalability}
        \label{fig:scalability}
    \end{subfigure}
    \begin{subfigure}[b]{0.32\columnwidth}
        \centering
        \includegraphics[width=\columnwidth]{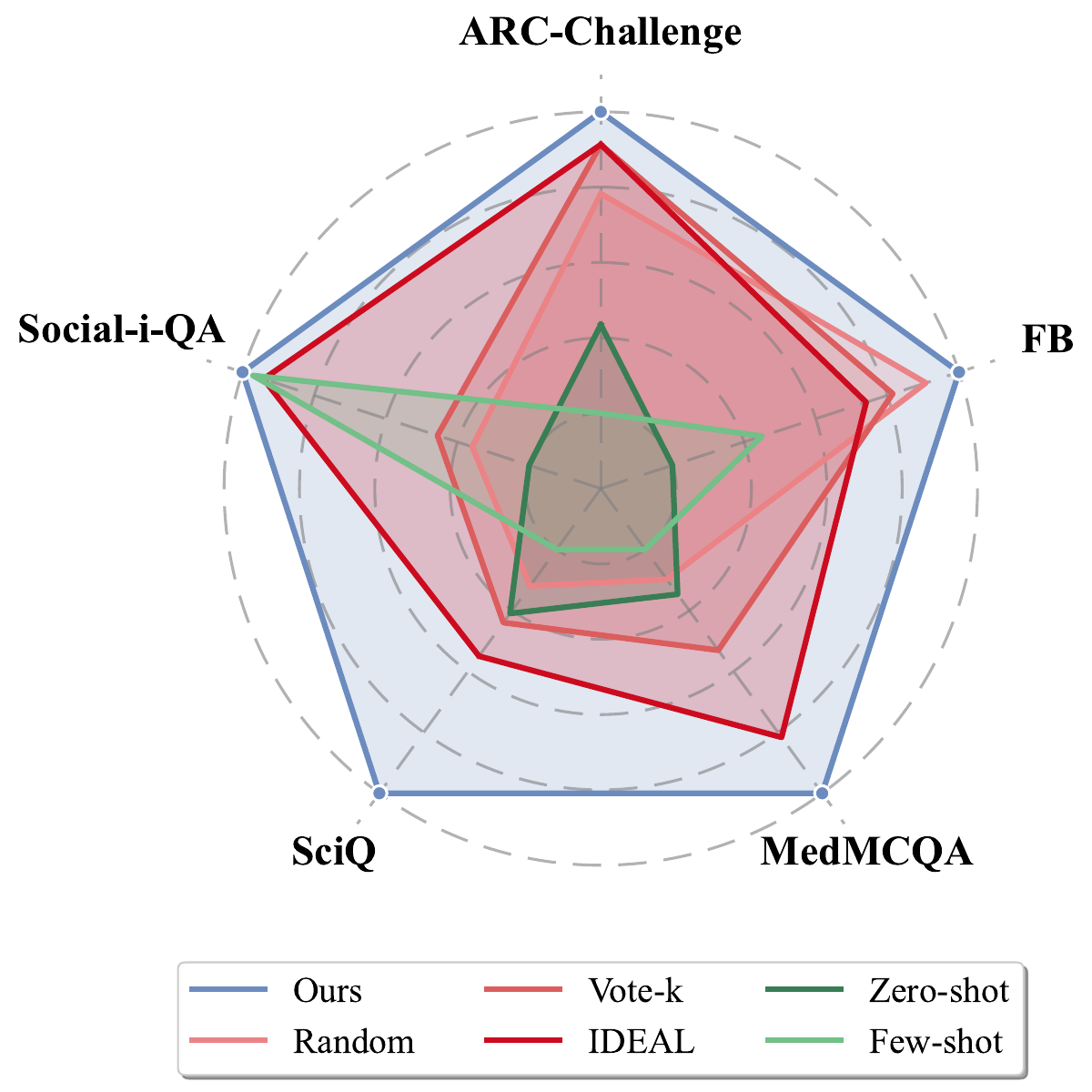}
        \caption{Ablation Study}
        \label{fig:ablation_study}
    \end{subfigure}
\caption{Detailed analysis of our proposed method.}
\label{fig:detailed_analysis}
\end{figure}

In this part, we conduct experiments on the Qwen-series LLMs ranging from 0.5B to 32B parameters in Figure~\ref{fig:different_llms}.
Additional results are presented in Figure~\ref{fig:addition-different-llms}.
Our findings indicate that few-shot prompting with cross-task examples consistently yields moderate improvements across all model scales, as these high-resource examples provide helpful knowledge that benefits low-resource ones.
Furthermore, \OURS outperforms these few-shot baselines via activation steering, which further demonstrates the effectiveness of our proposed method with different model sizes.

\subsubsection{The Scalability of \OURS}

To evaluate the scalability of our proposed method, we conduct experiments using different numbers of examples from the source task, as shown in Fugre~\ref{fig:scalability}.
Additional experiments are provided in Figure~\ref{fig:addition-scalability}.
We find that the performance of cross-task in-context learning initially improves with more examples, but eventually plateaus or even drops when too many examples are used.
This is partly due to the limited long-context capabilities of LLMs.
In addition, the number of demonstrations cannot scale indefinitely due to the maximum input length, which prevents the full utilization of large-scale high-resource data.
In contrast, our method shows a positive correlation between the number of examples and model performance.
With more demonstrations, our approach better isolates instance-level variations and extracts more general task-level features, leading to more effective cross-task transfer.
Importantly, since our approach does not rely on expanding the input context, it completely avoids the context window constraints. 
These results confirm the strong scalability of \OURS, especially in scenarios where abundant source data is available.

\subsubsection{The Efficiency of \OURS}

In this part, we discuss the computational efficiency of \OURS.
In more detail, our approach maintains the same time complexity as zero-shot prompting and is significantly lower than few-shot.
This is because we inject activations into the model's latent space during the forward pass without adding extra tokens to the input.
As a result, \OURS achieves effective cross-task transfer while preserving computational efficiency.

\subsubsection{Ablation Study}

Our approach introduces two key components: (1) Influential and diverse subset selection, and (2) {contrastive representation-enhanced activation steering}.
To verify the effectiveness of each component, we conduct ablation experiments using BoolQ as the source domain.
For the subset selection module, we compare it with three alternative strategies~(\ie random selection, Vote-k~(which focuses on diversity)~\cite{votek}, and IDEAL~(which emphasizes influence)~\cite{ideal}).
For the activation steering module, we compare it against standard zero-shot and few-shot prompting methods.
As shown in Figure~\ref{fig:ablation_study}, removing or replacing each component leads to a performance drop.
It indicates that both components are helpful.
Among them, both diversity and influence are essential for subset selection, since removing either of them results in performance drop.

%% file: tables/main_cd.tex
\begin{table*}[t]
    \centering
    \caption{Performance comparison in the cross-domain scenarios.}
        \resizebox{\columnwidth}{!}{
        \begin{tabular}{c|c|ccccccc|c}
            \toprule
             \multirow{2.5}{*}{\begin{tabular}[c]{@{}c@{}}\textbf{Target Domain}\end{tabular}} & \multirow{2.5}{*}{\begin{tabular}[c]{@{}c@{}}\textbf{Method}\end{tabular}} & \multicolumn{8}{c}{\textbf{Source Domain}} \\
            \cmidrule{3-10}
            & & ARC-Easy & AG-news & BoolQ & Com-QA & MNLI & QQP & SST2 & Average \\
            \midrule
            \multirow{4}{*}{\begin{tabular}[c]{@{}c@{}}\textbf{ARC-Challenge}\\ (Zs: 79.40) \end{tabular}} 
            & Few-shot Random & 81.00 & 77.60 & 79.80 & 77.60 & 78.00 & 78.40 & 79.00 & 79.08  \\
            & Few-shot TopK & 82.20 & 79.60 & \textbf{81.20} & 79.20 & 79.80 & 78.80 & 79.00 & 79.97  \\
            & Few-shot DPP  & 81.40 & 81.00 & 79.80 & 80.00 & 80.20 & 78.00 & 78.20 & 79.80  \\
            \cmidrule{2-10}
            & \cellcolor{gray!20}\OURS         & \cellcolor{gray!20}\textbf{83.80} & \cellcolor{gray!20}\textbf{81.20} & \cellcolor{gray!20}\textbf{81.20} & \cellcolor{gray!20}\textbf{81.60} & \cellcolor{gray!20}\textbf{82.40} & \cellcolor{gray!20}\textbf{81.20} & \cellcolor{gray!20}\textbf{82.40} & \cellcolor{gray!20}\textbf{81.97}  \\
            \cmidrule{1-10}
            \multirow{4}{*}{\begin{tabular}[c]{@{}c@{}}\textbf{Financial Phrasebank}\\ (Zs: 78.40) \end{tabular}} 
            & Few-shot Random & 85.40 & 72.00 & 81.20 & 79.60 & 70.20 & 78.20 & 75.40 & 77.43  \\
            & Few-shot TopK & 84.40 & 73.80 & 79.80 & 81.00 & 69.60 & 79.40 & 74.40 & 77.51  \\
            & Few-shot DPP  & 84.60 & 71.60 & 81.20 & 80.80 & 70.40 & 79.60 & 74.60 & 77.54  \\
            \cmidrule{2-10}
            & \cellcolor{gray!20}\OURS         & \cellcolor{gray!20}\textbf{86.40} & \cellcolor{gray!20}\textbf{81.20} & \cellcolor{gray!20}\textbf{83.40} & \cellcolor{gray!20}\textbf{82.20} & \cellcolor{gray!20}\textbf{80.20} & \cellcolor{gray!20}\textbf{81.20} & \cellcolor{gray!20}\textbf{84.00} & \cellcolor{gray!20}\textbf{82.66}   \\
            \cmidrule{1-10}
            \multirow{4}{*}{\begin{tabular}[c]{@{}c@{}}\textbf{MedMCQA}\\ (Zs: 53.80) \end{tabular}} 
            & Few-shot Random & 54.20 & 56.80 & 58.20 & 55.20 & 56.40 & 55.00 & 57.80 & 56.23  \\
            & Few-shot TopK & 57.40 & 55.40 & 59.40 & 54.80 & \textbf{57.80} & 55.20 & 58.20 & 56.89  \\
            & Few-shot DPP  & 59.00 & 57.80 & 59.60 & 52.60 & 56.20 & 56.20 & 57.80 & 57.03  \\
            \cmidrule{2-10}
            & \cellcolor{gray!20}\OURS         & \cellcolor{gray!20}\textbf{62.00} & \cellcolor{gray!20}\textbf{58.00} & \cellcolor{gray!20}\textbf{60.40} & \cellcolor{gray!20}\textbf{62.20} & \cellcolor{gray!20}\textbf{57.80} & \cellcolor{gray!20}\textbf{58.40} & \cellcolor{gray!20}\textbf{59.00} & \cellcolor{gray!20}\textbf{59.69}  \\
            \cmidrule{1-10}
            \multirow{4}{*}{\begin{tabular}[c]{@{}c@{}}\textbf{SciQ}\\ (Zs: 91.20) \end{tabular}}
            & Few-shot Random & 92.40 & 91.00 & 91.20 & 91.40 & 91.20 & 90.60 & 92.20 & 91.43   \\
            & Few-shot TopK & 92.20 & 90.80 & 91.80 & 91.40 & 91.00 & 92.40 & 91.40 & 91.57  \\
            & Few-shot DPP  & 93.00 & 89.60 & 91.20 & 92.80 & 91.20 & 91.00 & 91.00 & 91.40  \\
            \cmidrule{2-10}
            & \cellcolor{gray!20}\OURS         & \cellcolor{gray!20}\textbf{95.00} & \cellcolor{gray!20}\textbf{94.40} & \cellcolor{gray!20}\textbf{94.80} & \cellcolor{gray!20}\textbf{94.80} & \cellcolor{gray!20}\textbf{94.60} & \cellcolor{gray!20}\textbf{94.20} & \cellcolor{gray!20}\textbf{94.40} & \cellcolor{gray!20}\textbf{94.74}   \\
            \cmidrule{1-10}
            \multirow{4}{*}{\begin{tabular}[c]{@{}c@{}}\textbf{Social-i-QA}\\ (Zs: 69.40) \end{tabular}} 
            & Few-shot Random & 67.40 & 68.20 & 69.40 & 70.60 & 72.00 & 70.40 & 68.60 & 69.51  \\
            & Few-shot TopK & 69.60 & 69.20 & 69.60 & 70.00 & 69.80 & 70.00 & 69.60 & 69.69   \\
            & Few-shot DPP  & 71.80 & 69.60 & 68.00 & 69.40 & 70.20 & 69.40 & 69.60 & 69.71  \\
            \cmidrule{2-10}
            & \cellcolor{gray!20}\OURS         & \cellcolor{gray!20}\textbf{73.60} & \cellcolor{gray!20}\textbf{72.40} & \cellcolor{gray!20}\textbf{71.80} & \cellcolor{gray!20}\textbf{72.60} & \cellcolor{gray!20}\textbf{73.20} & \cellcolor{gray!20}\textbf{72.20} & \cellcolor{gray!20}\textbf{71.60} & \cellcolor{gray!20}\textbf{72.49}   \\
            \bottomrule
        \end{tabular}
        }
    \label{tab:main-cd}
\end{table*}

%% file: tables/name_cl.tex
\begin{table*}[t]
    \centering
    \small
    \caption{Performance comparison in the cross-lingual scenarios.}
        \resizebox{0.8\textwidth}{!}{
        \begin{tabular}{c|c|cccccc|c}
            \toprule
            \multirow{2.5}{*}{\begin{tabular}[c]{@{}c@{}}\textbf{Target} \\ \textbf{Language}\end{tabular}} & \multirow{2.5}{*}{\begin{tabular}[c]{@{}c@{}}\textbf{Method}\end{tabular}} & \multicolumn{7}{c}{\textbf{Source Language}} \\
            \cmidrule{3-9}
            & & de & en & es & fr & ja & zh & Average \\
            \midrule
            \multirow{4}{*}{\begin{tabular}[c]{@{}c@{}}\textbf{de}\\ (Zs: 87.40)\end{tabular}} 
            & Few-shot Random & - & 76.80 & 84.00 & 93.00 & 92.60 & 87.80 & 86.84 \\
            & Few-shot TopK   & - & 77.20 & 85.40 & 93.80 & 91.80 & 89.40 & 87.52 \\
            & Few-shot DPP    & - & 77.20 & 86.20 & 92.60 & 92.20 & 90.60 & 87.76 \\
            \cmidrule{2-9}
            & \cellcolor{gray!20}\OURS         & \cellcolor{gray!20}- & \cellcolor{gray!20}\textbf{89.80} & \cellcolor{gray!20}\textbf{90.80} & \cellcolor{gray!20}\textbf{95.80} & \cellcolor{gray!20}\textbf{93.20} & \cellcolor{gray!20}\textbf{91.80} & \cellcolor{gray!20}\textbf{92.28}  \\
            \cmidrule{1-9}
            \multirow{4}{*}{\begin{tabular}[c]{@{}c@{}}\textbf{en}\\ (Zs: 76.00)\end{tabular}} 
            & Few-shot Random & 77.40 & - & 42.80 & 58.00 & 76.80 & 41.80 & 59.36  \\
            & Few-shot TopK   & 76.80 & - & 51.00 & 62.40 & 79.60 & 38.00 & 61.56 \\
            & Few-shot DPP    & 76.80 & - & 58.80 & 68.80 & 80.00 & 38.00 & 64.48 \\
            \cmidrule{2-9}
            & \cellcolor{gray!20}\OURS         & \cellcolor{gray!20}\textbf{90.60} & \cellcolor{gray!20}- & \cellcolor{gray!20}\textbf{86.00} & \cellcolor{gray!20}\textbf{91.20} & \cellcolor{gray!20}\textbf{90.40} & \cellcolor{gray!20}\textbf{87.00} & \cellcolor{gray!20}\textbf{89.04}  \\
            \cmidrule{1-9}
            \multirow{4}{*}{\begin{tabular}[c]{@{}c@{}}\textbf{es}\\ (Zs: 91.60)\end{tabular}} 
            & Few-shot Random & 93.80 & 93.80 & - & 93.60 & 91.60 & 92.20 & 93.00 \\
            & Few-shot TopK  & 94.00 & 94.00 & - & 94.20 & 93.00 & 93.20 & 93.68  \\
            & Few-shot DPP   & 94.00 & 94.20 & - & 93.60 & 93.40 & 93.40 & 93.72 \\
            \cmidrule{2-9}
            & \cellcolor{gray!20}\OURS         & \cellcolor{gray!20}\textbf{95.20} & \cellcolor{gray!20}\textbf{95.60} & \cellcolor{gray!20}- & \cellcolor{gray!20}\textbf{95.20} & \cellcolor{gray!20}\textbf{94.60} & \cellcolor{gray!20}\textbf{93.80} & \cellcolor{gray!20}\textbf{94.88} \\
            \cmidrule{1-9}
            \multirow{4}{*}{\begin{tabular}[c]{@{}c@{}}\textbf{fr}\\ (Zs: 89.60)\end{tabular}} 
            & Few-shot Random & 91.80 & 65.60 & 84.00 & - & 91.80 & 84.20 & 83.48  \\
            & Few-shot TopK   & 91.40 & 65.40 & 86.80 & - & 92.00 & 82.80 & 83.68  \\
            & Few-shot DPP    & 91.40 & 69.40 & 89.40 & - & 92.20 & 84.60 & 85.40 \\
            \cmidrule{2-9}
            & \cellcolor{gray!20}\OURS         & \cellcolor{gray!20}9\textbf{5.00} & \cellcolor{gray!20}\textbf{95.20} & \cellcolor{gray!20}\textbf{95.80} & \cellcolor{gray!20}- & \cellcolor{gray!20}\textbf{95.40} & \cellcolor{gray!20}\textbf{94.60} & \cellcolor{gray!20}\textbf{95.20} \\
            \cmidrule{1-9}
            \multirow{4}{*}{\begin{tabular}[c]{@{}c@{}}\textbf{ja}\\ (Zs: 40.60)\end{tabular}} 
            & Few-shot Random & 49.00 & 27.80 & 44.60 & 45.00 & - & 36.00 & 40.48 \\
            & Few-shot TopK   & 49.40 & 26.00 & 45.40 & 45.60 & - & 36.60 & 40.60 \\
            & Few-shot DPP    & 50.40 & 25.20 & 47.40 & 46.00 & - & 36.20 & 41.04 \\
            \cmidrule{2-9}
            & \cellcolor{gray!20}\OURS         & \cellcolor{gray!20}\textbf{53.20} & \cellcolor{gray!20}\textbf{47.00} & \cellcolor{gray!20}\textbf{47.60} & \cellcolor{gray!20}\textbf{47.00} & \cellcolor{gray!20}- & \cellcolor{gray!20}\textbf{45.80} & \cellcolor{gray!20}\textbf{48.12} \\
            \cmidrule{1-9}
            \multirow{4}{*}{\begin{tabular}[c]{@{}c@{}}\textbf{zh}\\ (Zs: 30.80)\end{tabular}} 
            & Few-shot Random & 49.00 & 35.00 & 33.60 & 32.00 & 37.20 & - & 37.36 \\
            & Few-shot TopK   & 49.20 & 36.60 & 35.80 & 33.20 & 37.00 & - & 38.36 \\
            & Few-shot DPP    & 51.60 & 35.80 & 37.00 & 31.40 & 37.80 & - & 38.72 \\
            \cmidrule{2-9}
            & \cellcolor{gray!20}\OURS         & \cellcolor{gray!20}\textbf{53.20} & \cellcolor{gray!20}\textbf{42.80} & \cellcolor{gray!20}\textbf{42.20} & \cellcolor{gray!20}\textbf{43.20} & \cellcolor{gray!20}\textbf{43.00} & \cellcolor{gray!20}- & \cellcolor{gray!20}\textbf{44.88} \\
            \bottomrule
        \end{tabular}
        }
    \label{tab:main-cl}
\end{table*}

%% file: sections/conclusion.tex
\section{Conclusion}

In this work, we explored the potential of achieving cross-task transfer in LLMs via latent space steering. 
Through empirical analysis, we observed that in-context examples consistently induce enhanced activation patterns in the model’s latent space across different tasks.
Building on this insight, we proposed \OURS, a novel framework that steers activations from high-resource tasks to enhance the performance of low-resource ones, enabling effective cross-task transfer.
A key advantage of our approach lies in its robustness, scalability, and computational efficiency, for it leverages pre-computed activations during inference, without updating parameters or increasing input length.
Extensive experiments on both cross-domain and cross-lingual transfer scenarios demonstrated the effectiveness of our proposed method.

%% file: sections/appendix/algorithm.tex
\input{algorithms/qua_algo}
\input{algorithms/selection_algo}

%% file: algorithms/qua_algo.tex
\begin{algorithm*}[t]
\centering
\caption{$\operatorname{Influence \ and \ Diversity-driven \ Sample \ Quantification}(\mathcal{G}, S_{\text{selected}}, N_i(\cdot), v, k, \alpha, \gamma)$}
\label{alg:quantification}
\begin{minipage}{1\linewidth} 
\small
\begin{algorithmic}
    \Inputs{
        Sample directed graph $\mathcal{G} = (\mathbf{V}, \mathbf{E},\mathbf{P})$, $i$-hop neighibor function $N_i(\cdot)$, Current selected sample subset $S_{\text{selected}}$, Sample node ${v}$, Neighborhood depth $k$, Hop-based decay factor $\alpha$, balance hyper-parameter between diversity and influence $\gamma$.
    }
    \Initialize{
        $D_v=0$, $I_v=0$, $S_{\text{active}} \gets {v}$, $S_{\text{visited}} \gets \emptyset$
    }
    \While {$S_{\text{active}} \neq \emptyset$} \Comment{Influencial Calculation}
        \State Choose a sample node $u \in S_{\text{active}}$
        \For{each neighbor $w \in N_1(u)$}
            \State Select edge $(u,w)$ with probability $p(u, w)$
            \If{edge $(u,w)$ is selected \textbf{and} $w \notin S_{\text{visited}}$}
                \State $S_{\text{active}} \gets S_{\text{active}} \cup w$, $S_{\text{visited}} \gets S_{\text{visited}} \cup w$
            \EndIf
        \EndFor
        \State $S_{\text{active}} \gets S_{\text{active}} \setminus u$
    \EndWhile
    \State $I_v = |S_{\text{visited}}|$    
    \For{$i = 1$ to $k$} \Comment{Diversity Calculation}
        \State Search $i$-hop neighbors of sample node $v$: $N_i({v})$;
        \State Compute overlap between $i$-hop neighbors and $S_{\text{selected}}$: $o_i \gets |N_i({v}) \cap S_{\text{selected}}|$
        \State $D_v \gets D_v - \alpha^i \cdot o_i$
    \EndFor
    \State \Return Sample node ${v}$ evaluation function: $ f_{\mathcal{G}}({v}) \gets I_v + \gamma \cdot D_v$
\end{algorithmic}
\end{minipage}
\end{algorithm*}

%% file: algorithms/selection_algo.tex
\begin{algorithm*}[t]
\centering
\caption{$\operatorname{Iterative \ Greedy \ Graph \ Search}(\mathcal{G}, S, n)$}
\label{alg:selection}
\begin{minipage}{1\linewidth} 
\small
\begin{algorithmic}
    \Inputs{
        Sample directed graph $\mathcal{G} = (\mathbf{V}, \mathbf{E},\mathbf{P})$, Initial sample subset $S_0$, Selected sample subset size $n$.
    }
    \Initialize{
        $\mathcal{S}_0 \to \emptyset, i=0$,
        Sample node evaluation function $f_\mathcal{G}: \mathbf{V} \mapsto \mathbb R$ based on Algorithm~\ref{alg:quantification} 
    }
    \While {$i < n$}
        \State ${v}^{*} \gets \mathop{\arg\max} \limits_{{v} \in \mathbf{V} \setminus \mathcal{S}_{i}} f_{\mathcal{G}}({v})$
        \State $\mathcal{S}_{i+1} \gets \mathcal{S}_{i} \cup {v}^{*}$
        \State $i \gets i+1$ 
    \EndWhile
    \State \Return{$\mathcal{S}_{n}$.} 
\end{algorithmic}
\end{minipage}
\end{algorithm*}

%% file: sections/appendix/experiments.tex
\section{Additional Experiments}
\label{app:additional-experiments}

\begin{figure*}[t]
    \centering
    \begin{subfigure}[b]{0.32\textwidth}
        \centering
        \includegraphics[width=\textwidth]{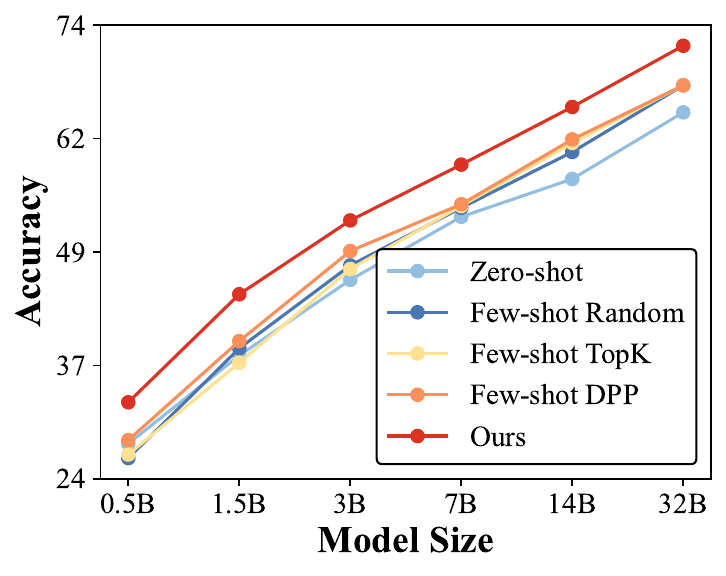}
        \caption{Source Task: MNLI Target Task: MedMCQA}
        \label{subfig:different-llms1}
    \end{subfigure}
    \begin{subfigure}[b]{0.32\textwidth}
        \centering
        \includegraphics[width=\textwidth]{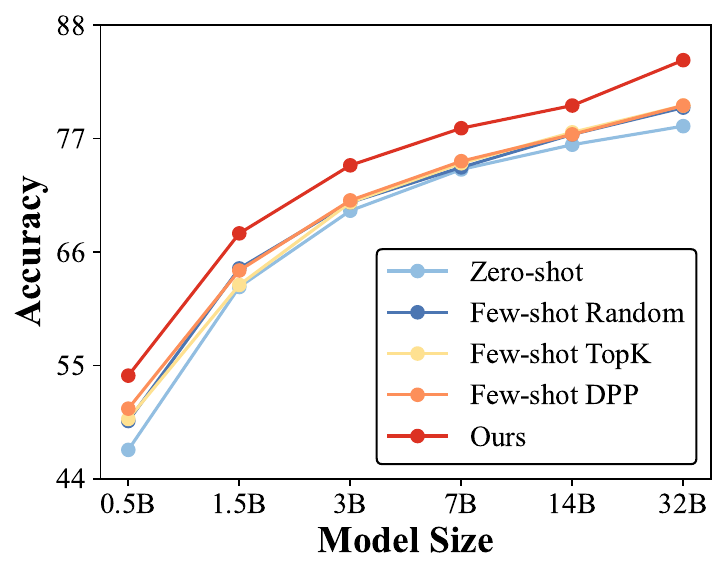}
        \caption{Source Task: Com-QA Target Task: Social-i-QA}
        \label{subfig:different-llms2}
    \end{subfigure}
    \begin{subfigure}[b]{0.32\textwidth}
        \centering
        \includegraphics[width=\textwidth]{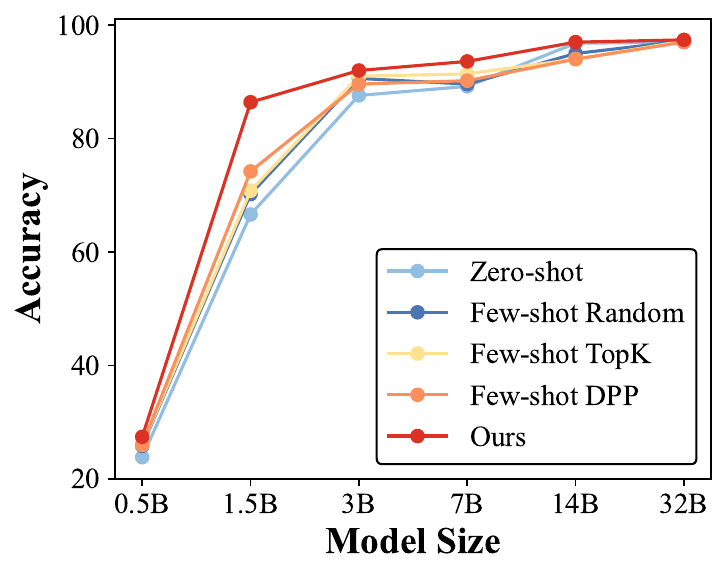}
        \caption{Source Task: SST2 Target Task: Financial Phrasebank}
        \label{subfig:different-llms3}
    \end{subfigure}
\caption{Different Model Sizes}
\label{fig:addition-different-llms}
\end{figure*}

\begin{figure*}[t]
    \centering
    \begin{subfigure}[b]{0.32\textwidth}
        \centering
        \includegraphics[width=\textwidth]{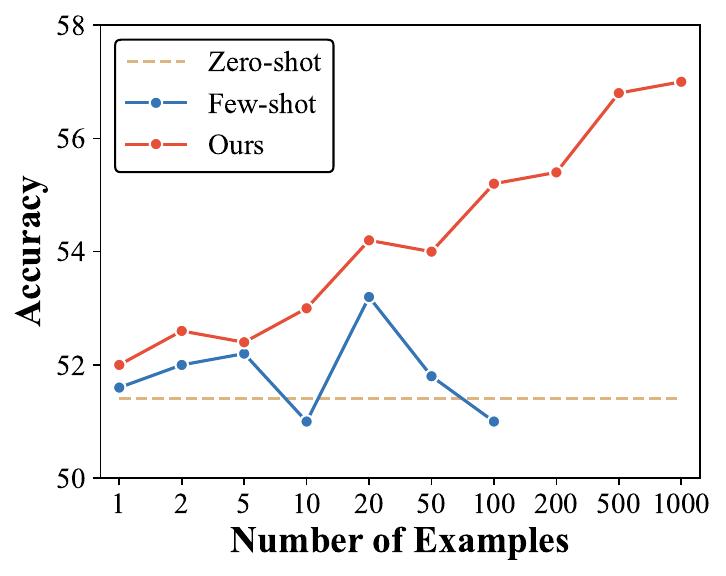}
        \caption{Source Task: MNLI, Target Task: MedMCQA}
        \label{subfig:scalability-1}
    \end{subfigure}
    \begin{subfigure}[b]{0.32\textwidth}
        \centering
        \includegraphics[width=\textwidth]{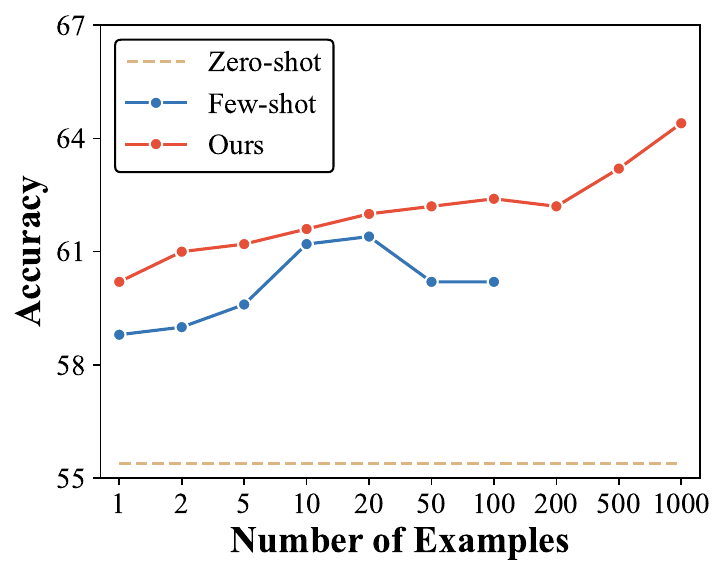}
        \caption{Source Task: Com-QA, Target Task: Social-i-QA}
        \label{subfig:scalability-2}
    \end{subfigure}
    \begin{subfigure}[b]{0.32\textwidth}
        \centering
        \includegraphics[width=\textwidth]{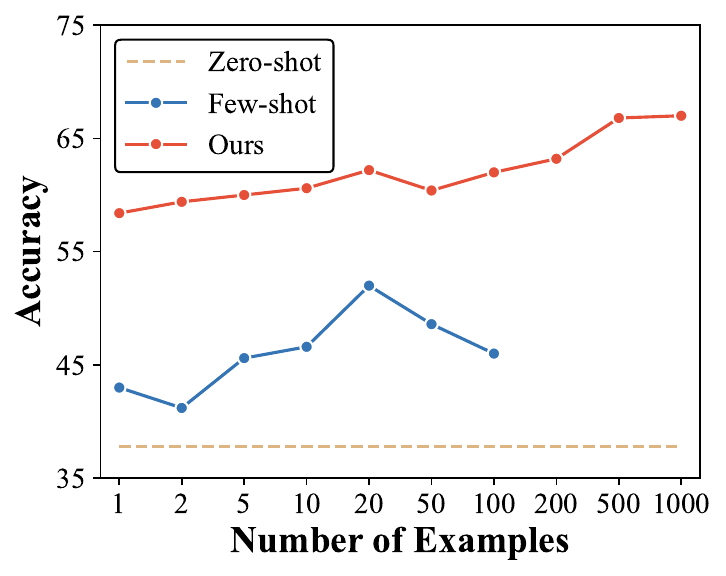}
        \caption{Source Task: SST2, Target Task: Financial Phrasebank}
        \label{subfig:scalability-3}
    \end{subfigure}
\caption{Scalability}
\label{fig:addition-scalability}
\end{figure*}

\subsection{Experiment on Different LLMs}
\label{app-sub:different-llm}

\input{tables/llama_cd}
\input{tables/qwen_cd}

To further validate the effectiveness of our proposed method across different LLMs, in this part, we conduct experiments on both base models (\ie Llama3.1-8B and Qwen-2.5-7B) and their instruction-tuned counterparts (\ie Llama3.1-8B-Instruct and Qwen2.5-7B-Instruct).
As shown in Table~\ref{tab:llama-cd} and Table~\ref{tab:main-qwen-cd}, our method consistently outperforms all baselines across different LLMs.

\subsection{Experiments on Generation Tasks}

\input{tables/generation}

To evaluate the generalizability of our proposed method beyond classification tasks, we conduct experiments on generation tasks.
We use MATH~\cite{MATH-Dataset} as the source task, and GPQA~\cite{GPQA-Dataset} and LiveCodeBench~\cite{LiveCodeBench-Dataset} as the target tasks.
As presented in Table~\ref{tab:generation_task}, adding cross-task examples in generation tasks leads to performance degradation.
In contrast, \OURS achieves the best performance by leveraging activation steering in the latent space of LLMs, which demonstrates the robustness and effectiveness of our method.

\subsection{Hyper-parameter Analysis}

\OURS includes a few hyperparameters to tune.
In this section, we report the tuning results of several hyperparameters.

\begin{figure*}[t]
    \centering
    \begin{subfigure}[b]{0.32\textwidth}
        \centering
        \includegraphics[width=\textwidth]{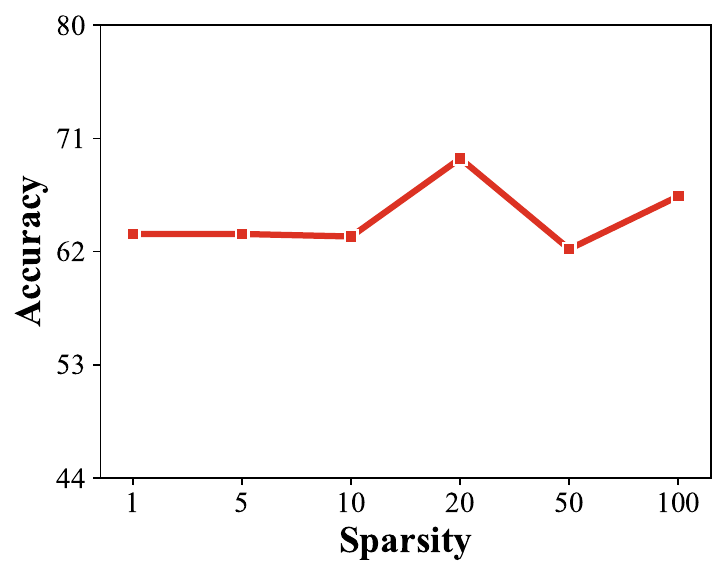}
        \caption{Graph sparsity parameter}
        \label{fig:beta}
    \end{subfigure}
    \begin{subfigure}[b]{0.32\textwidth}
        \centering
        \includegraphics[width=\textwidth]{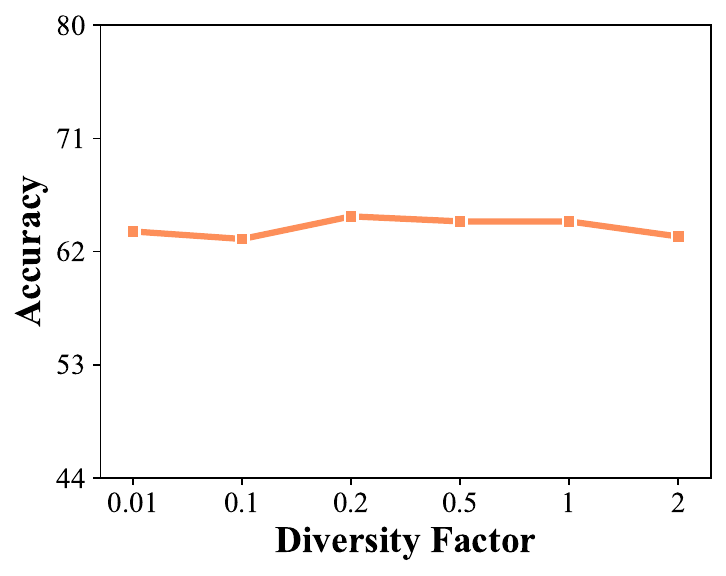}
        \caption{Hop-based decay factor}
        \label{fig:alpha}
    \end{subfigure}
    \begin{subfigure}[b]{0.32\textwidth}
        \centering
        \includegraphics[width=\textwidth]{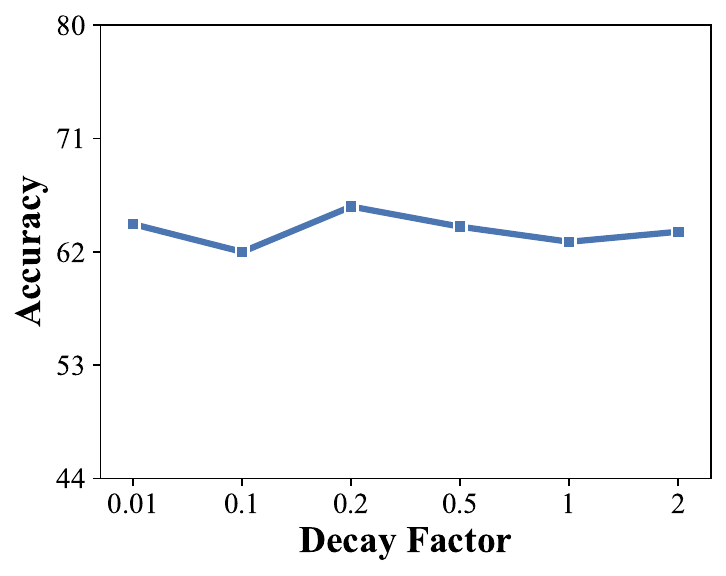}
        \caption{Trade-off parameter}
        \label{fig:gamma}
    \end{subfigure}
\caption{Hyperparameter analysis of the subset selection method.}
\label{fig:subset-hyper}
\end{figure*}

For the influential and diverse subset selection method, we explore the parameter $\beta$ controlling the graph sparsity, the hop-based decay factor $\alpha$, and the trade-off parameter $\gamma$ between diversity and influence.
The results are shown in Figure~\ref{fig:beta}, Figure~\ref{fig:alpha}, Figure~\ref{fig:gamma}, respectively.
These hyperparameters have only a minor impact on overall performance.

\begin{figure*}[t]
    \centering
    \begin{subfigure}[b]{0.48\textwidth}
        \centering
        \includegraphics[width=\textwidth]{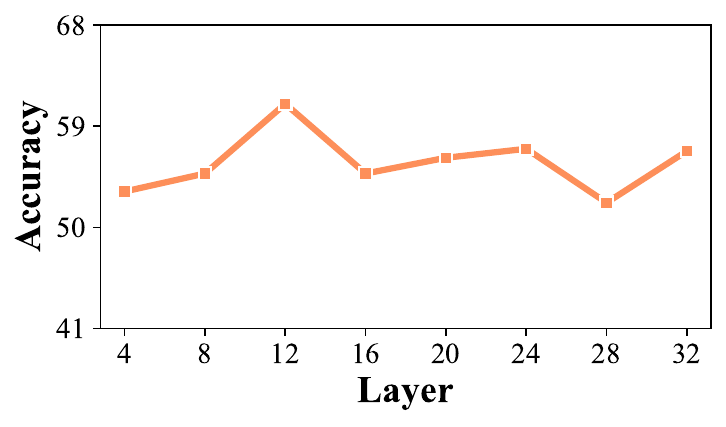}
        \caption{Layer selection}
        \label{fig:layer-selection}
    \end{subfigure}
    \begin{subfigure}[b]{0.48\textwidth}
        \centering
        \includegraphics[width=\textwidth]{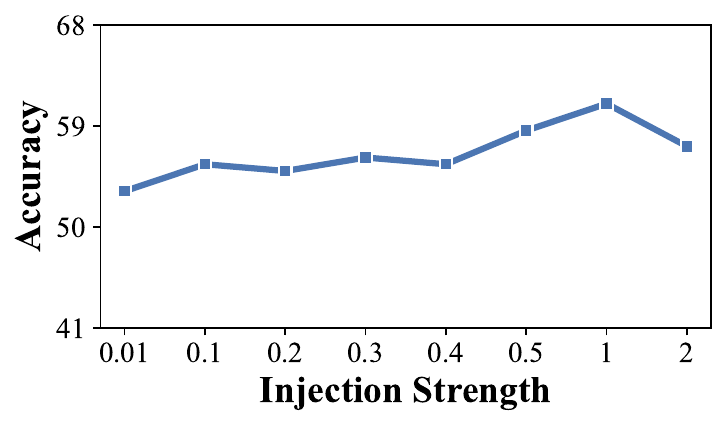}
        \caption{Injection Strength}
        \label{fig:injection-strength}
    \end{subfigure}
\caption{Hyperparameter analysis of the activation selection approach.}
\label{fig:activation-steering-hyper}
\end{figure*}

For the activation steering approach, we explore the effect of layer selection in Figure~\ref{fig:layer-selection}, injection strength in Figure~\ref{fig:injection-strength}, and injection position in Table~\ref{tab:injection_position}.
Regarding layer selection, we observe better performance when activations are injected into middle layers.
This may be because middle layers capture richer semantic features in LLMs.
For injection strength, we find that the performance is optimal when the injection strength is set to 1.
If the injection strength is too small, the model cannot effectively incorporate information from the high-resource task.
Conversely, if it is too large, the injected activations can disrupt the original input semantics, leading to performance degradation.
Finally, for the injection position, we find that injecting activations at the last token leads to the best performance, for it allows the model to directly influence the generation process without interfering with the representations of earlier tokens.

\input{tables/injection_position}

%% file: tables/llama_cd.tex
\begin{table*}[t]
    \centering
    \caption{Experiments on cross-domain scenarios using Llama3.1-8B and Llama3.1-8B-Instruct.}
    \resizebox{\columnwidth}{!}{
    \begin{tabular}{c|c|c|ccccccc|c}
        \toprule
        \multirow{2.5}{*}{\begin{tabular}[c]{@{}c@{}}\textbf{Model}\end{tabular}} & \multirow{2.5}{*}{\begin{tabular}[c]{@{}c@{}}\textbf{Target Task}\end{tabular}} & \multirow{2.5}{*}{\begin{tabular}[c]{@{}c@{}}\textbf{Method}\end{tabular}} & \multicolumn{8}{c}{\textbf{Source Task}} \\
        \cmidrule{4-11}
        & & & ARC-Easy & AG-news & BoolQ & Com-QA & MNLI & QQP & SST2 & Average \\
        \midrule
        \multirow{25}{*}{\begin{tabular}[c]{@{}c@{}}\rotatebox{90}{\textbf{Llama3.1-8B}}\end{tabular}} & \multirow{4}{*}{\begin{tabular}[c]{@{}c@{}}\textbf{ARC-Challenge}\\ (Zs: 71.80) \end{tabular}} 
        & Few-shot Random & 72.40 & 68.00 & 72.40 & 70.60 & 64.60 & 67.20 & 70.00 & 69.31 \\
        & & Few-shot TopK & 74.20 & 69.20 & 69.80 & 71.20 & 66.20 & 67.40 & 68.80 & 69.54 \\
        & & Few-shot DPP  & 75.20 & 66.40 & 71.00 & 71.40 & 67.80 & 67.00 & 69.40 & 69.74 \\
        \cmidrule{3-11}
        & & \cellcolor{gray!20}\OURS         & \cellcolor{gray!20}\textbf{75.80} & \cellcolor{gray!20}\textbf{76.40} & \cellcolor{gray!20}\textbf{75.60} & \cellcolor{gray!20}\textbf{77.20} & \cellcolor{gray!20}\textbf{76.20} & \cellcolor{gray!20}\textbf{75.60} & \cellcolor{gray!20}\textbf{76.20} & \cellcolor{gray!20}\textbf{76.14} \\
        \cmidrule{2-11}
        & \multirow{4}{*}{\begin{tabular}[c]{@{}c@{}}\textbf{Financial Phrasebank}\\ (Zs: 37.80) \end{tabular}} 
        & Few-shot Random & 44.80 & 48.20 & 46.40 & 48.80 & 56.80          & 48.60 & 46.00 & 48.51   \\
        & & Few-shot TopK & 42.00 & 47.20 & 46.00 & 47.00 & 61.60          & 52.40 & 44.20 & 48.63   \\
        & & Few-shot DPP  & 40.80 & 47.60 & 46.80 & 49.60 & \textbf{61.80} & 49.80 & 45.80 & 48.89   \\
        \cmidrule{3-11}
        & & \cellcolor{gray!20}\OURS         & \cellcolor{gray!20}\textbf{54.60} & \cellcolor{gray!20}\textbf{60.40} & \cellcolor{gray!20}\textbf{59.40} & \cellcolor{gray!20}\textbf{53.40} & \cellcolor{gray!20}60.00 & \cellcolor{gray!20}\textbf{64.40} & \cellcolor{gray!20}\textbf{54.60} & \cellcolor{gray!20}\textbf{58.11}   \\
        \cmidrule{2-11}
        & \multirow{4}{*}{\begin{tabular}[c]{@{}c@{}}\textbf{MedMCQA}\\ (Zs: 49.40) \end{tabular}} 
        & Few-shot Random & 47.00 & 50.00 & 46.80 & 47.60 & 53.20 & 46.60 & 49.60 & 48.69   \\
        & & Few-shot TopK & 47.80 & 49.00 & 47.80 & 47.40 & 52.80 & 47.40 & 48.60 & 48.69   \\
        & & Few-shot DPP  & 49.00 & 49.80 & 50.20 & 47.40 & 54.00 & 47.20 & 54.60 & 50.31   \\
        \cmidrule{3-11}
        & & \cellcolor{gray!20}\OURS         & \cellcolor{gray!20}\textbf{57.20} & \cellcolor{gray!20}\textbf{54.80} & \cellcolor{gray!20}\textbf{55.10} & \cellcolor{gray!20}\textbf{55.80} & \cellcolor{gray!20}\textbf{56.00} & \cellcolor{gray!20}\textbf{55.20} & \cellcolor{gray!20}\textbf{56.20} & \cellcolor{gray!20}\textbf{55.76}   \\
        \cmidrule{2-11}
        & \multirow{4}{*}{\begin{tabular}[c]{@{}c@{}}\textbf{SciQ}\\ (Zs: 84.40) \end{tabular}} 
        & Few-shot Random & 88.20 & 84.20 & 86.60 & 87.80 & 80.00 & 81.80 & 83.80 & 84.63   \\
        & & Few-shot TopK & 87.60 & 82.20 & 85.80 & 87.00 & 83.20 & 81.60 & 86.60 & 84.86   \\
        & & Few-shot DPP  & 87.40 & 86.40 & 87.00 & 87.80 & 82.60 & 82.00 & 83.60 & 85.26   \\
        \cmidrule{3-11}
        & & \cellcolor{gray!20}\OURS         & \cellcolor{gray!20}\textbf{89.60} & \cellcolor{gray!20}\textbf{89.00} & \cellcolor{gray!20}\textbf{91.00} & \cellcolor{gray!20}\textbf{90.00} & \cellcolor{gray!20}\textbf{89.00} & \cellcolor{gray!20}\textbf{89.80} & \cellcolor{gray!20}\textbf{89.60} & \cellcolor{gray!20}\textbf{89.71}   \\
        \cmidrule{2-11}
        & \multirow{4}{*}{\begin{tabular}[c]{@{}c@{}}\textbf{Social-i-QA}\\ (Zs: 55.40) \end{tabular}} 
        & Few-shot Random & 60.40 & 58.00 & 59.60 & 62.40          & 59.00 & 57.20 & 57.40 & 59.14   \\
        & & Few-shot TopK & 62.80 & 60.80 & 58.80 & \textbf{62.80} & 58.60 & 56.80 & 56.00 & 59.51  \\
        & & Few-shot DPP  & 60.20 & 61.00 & 62.00 & 60.00          & 60.00 & 59.60 & 56.40 & 59.89  \\
        \cmidrule{3-11}
        & & \cellcolor{gray!20}\OURS         & \cellcolor{gray!20}\textbf{65.60} & \cellcolor{gray!20}\textbf{63.20} & \cellcolor{gray!20}\textbf{62.20} & \cellcolor{gray!20}\textbf{62.40} & \cellcolor{gray!20}\textbf{63.40} & \cellcolor{gray!20}\textbf{65.60} & \cellcolor{gray!20}\textbf{63.60} & \cellcolor{gray!20}\textbf{63.71}  \\
        \midrule
        \multirow{25}{*}{\begin{tabular}[c]{@{}c@{}}\rotatebox{90}{\textbf{Llama3.1-8B-Instruct}}\end{tabular}} & \multirow{4}{*}{\begin{tabular}[c]{@{}c@{}}\textbf{ARC-Challenge}\\ (Zs: 79.40) \end{tabular}} 
        & Few-shot Random & 81.00 & 77.60 & 79.80 & 77.60 & 78.00 & 78.40 & 79.00 & 79.08  \\
        & & Few-shot TopK & 82.20 & 79.60 & \textbf{81.20} & 79.20 & 79.80 & 78.80 & 79.00 & 79.97  \\
        & & Few-shot DPP  & 81.40 & 81.00 & 79.80 & 80.00 & 80.20 & 78.00 & 78.20 & 79.80  \\
        \cmidrule{3-11}
        & & \cellcolor{gray!20}\OURS         & \cellcolor{gray!20}\textbf{83.80} & \cellcolor{gray!20}\textbf{81.20} & \cellcolor{gray!20}\textbf{81.20} & \cellcolor{gray!20}\textbf{81.60} & \cellcolor{gray!20}\textbf{82.40} & \cellcolor{gray!20}\textbf{81.20} & \cellcolor{gray!20}\textbf{82.40} & \cellcolor{gray!20}\textbf{81.97}  \\
        \cmidrule{2-11}
        & \multirow{4}{*}{\begin{tabular}[c]{@{}c@{}}\textbf{Financial Phrasebank}\\ (Zs: 78.40) \end{tabular}} 
        & Few-shot Random & 85.40 & 72.00 & 81.20 & 79.60 & 70.20 & 78.20 & 75.40 & 77.43  \\
        & & Few-shot TopK & 84.40 & 73.80 & 79.80 & 81.00 & 69.60 & 79.40 & 74.40 & 77.51  \\
        & & Few-shot DPP  & 84.60 & 71.60 & 81.20 & 80.80 & 70.40 & 79.60 & 74.60 & 77.54  \\
        \cmidrule{3-11}
        & & \cellcolor{gray!20}\OURS         & \cellcolor{gray!20}\textbf{86.40} & \cellcolor{gray!20}\textbf{81.20} & \cellcolor{gray!20}\textbf{83.40} & \cellcolor{gray!20}\textbf{82.20} & \cellcolor{gray!20}\textbf{80.20} & \cellcolor{gray!20}\textbf{81.20} & \cellcolor{gray!20}\textbf{84.00} & \cellcolor{gray!20}\textbf{82.66}   \\
        \cmidrule{2-11}
        & \multirow{4}{*}{\begin{tabular}[c]{@{}c@{}}\textbf{MedMCQA}\\ (Zs: 53.80) \end{tabular}} 
        & Few-shot Random & 54.20 & 56.80 & 58.20 & 55.20 & 56.40 & 55.00 & 57.80 & 56.23  \\
        & & Few-shot TopK & 57.40 & 55.40 & 59.40 & 54.80 & \textbf{57.80} & 55.20 & 58.20 & 56.89  \\
        & & Few-shot DPP  & 59.00 & 57.80 & 59.60 & 52.60 & 56.20 & 56.20 & 57.80 & 57.03  \\
        \cmidrule{3-11}
        & & \cellcolor{gray!20}\OURS         & \cellcolor{gray!20}\textbf{62.00} & \cellcolor{gray!20}\textbf{58.00} & \cellcolor{gray!20}\textbf{60.40} & \cellcolor{gray!20}\textbf{62.20} & \cellcolor{gray!20}\textbf{57.80} & \cellcolor{gray!20}\textbf{58.40} & \cellcolor{gray!20}\textbf{59.00} & \cellcolor{gray!20}\textbf{59.69}  \\
        \cmidrule{2-11}
        & \multirow{4}{*}{\begin{tabular}[c]{@{}c@{}}\textbf{SciQ}\\ (Zs: 91.20) \end{tabular}}
        & Few-shot Random & 92.40 & 91.00 & 91.20 & 91.40 & 91.20 & 90.60 & 92.20 & 91.43   \\
        & & Few-shot TopK & 92.20 & 90.80 & 91.80 & 91.40 & 91.00 & 92.40 & 91.40 & 91.57  \\
        & & Few-shot DPP  & 93.00 & 89.60 & 91.20 & 92.80 & 91.20 & 91.00 & 91.00 & 91.40  \\
        \cmidrule{3-11}
        & & \cellcolor{gray!20}\OURS         & \cellcolor{gray!20}\textbf{95.00} & \cellcolor{gray!20}\textbf{94.40} & \cellcolor{gray!20}\textbf{94.80} & \cellcolor{gray!20}\textbf{94.80} & \cellcolor{gray!20}\textbf{94.60} & \cellcolor{gray!20}\textbf{94.20} & \cellcolor{gray!20}\textbf{94.40} & \cellcolor{gray!20}\textbf{94.74}   \\
        \cmidrule{2-11}
        & \multirow{4}{*}{\begin{tabular}[c]{@{}c@{}}\textbf{Social-i-QA}\\ (Zs: 69.40) \end{tabular}} 
        & Few-shot Random & 67.40 & 68.20 & 69.40 & 70.60 & 72.00 & 70.40 & 68.60 & 69.51  \\
        & & Few-shot TopK & 69.60 & 69.20 & 69.60 & 70.00 & 69.80 & 70.00 & 69.60 & 69.69   \\
        & & Few-shot DPP  & 71.80 & 69.60 & 68.00 & 69.40 & 70.20 & 69.40 & 69.60 & 69.71  \\
        \cmidrule{3-11}
        & & \cellcolor{gray!20}\OURS         & \cellcolor{gray!20}\textbf{73.60} & \cellcolor{gray!20}\textbf{72.40} & \cellcolor{gray!20}\textbf{71.80} & \cellcolor{gray!20}\textbf{72.60} & \cellcolor{gray!20}\textbf{73.20} & \cellcolor{gray!20}\textbf{72.20} & \cellcolor{gray!20}\textbf{71.60} & \cellcolor{gray!20}\textbf{72.49}   \\
        \bottomrule
    \end{tabular}
    }
    \label{tab:llama-cd}
\end{table*}

%% file: tables/qwen_cd.tex
\begin{table*}[t]
    \centering
    \large
    \caption{Experiments on cross-domain scenarios using Qwen2.5-7B and Qwen2.5-7B-Instruct.}
    \resizebox{\columnwidth}{!}{
    \begin{tabular}{c|c|c|ccccccc|c}
        \toprule
        \multirow{2.5}{*}{\begin{tabular}[c]{@{}c@{}}\textbf{Model}\end{tabular}} & \multirow{2.5}{*}{\begin{tabular}[c]{@{}c@{}}\textbf{Target Task}\end{tabular}} & \multirow{2.5}{*}{\begin{tabular}[c]{@{}c@{}}\textbf{Method}\end{tabular}} & \multicolumn{8}{c}{\textbf{Source Task}} \\
        \cmidrule{4-11}
        & & & ARC-Easy & AG-news & BoolQ & Com-QA & MNLI & QQP & SST2 & Average \\
        \midrule
        \multirow{25}{*}{\begin{tabular}[c]{@{}c@{}}\rotatebox{90}{\textbf{Qwen2.5-7B}}\end{tabular}} & \multirow{4}{*}{\begin{tabular}[c]{@{}c@{}}\textbf{ARC-Challenge}\\ (Zs: 84.80) \end{tabular}} & Few-shot Random   & 89.80 & 89.60 & 89.20 & 89.60 & 89.40 & 89.20 & 89.80 & 89.51 \\
        & & Few-shot TopK & 90.40 & 89.40 & 89.00 & 89.60 & 89.60 & 89.60 & 89.60 & 89.60 \\
        & & Few-shot DPP  & 90.20 & 89.60 & 89.00 & 89.80 & \textbf{90.00} & 89.00 & 89.60 & 89.60   \\
        \cmidrule{3-11}
        & & \cellcolor{gray!20}\OURS         & \cellcolor{gray!20}\textbf{93.40} & \cellcolor{gray!20}\textbf{91.20} & \cellcolor{gray!20}\textbf{92.40} & \cellcolor{gray!20}\textbf{90.80} & \cellcolor{gray!20}\textbf{90.00} & \cellcolor{gray!20}\textbf{91.80} & \cellcolor{gray!20}\textbf{90.80} & \cellcolor{gray!20}\textbf{91.49}  \\
        \cmidrule{2-11}
        & \multirow{4}{*}{\begin{tabular}[c]{@{}c@{}}\textbf{Financial Phrasebank}\\ (Zs: 89.20) \end{tabular}} 
        & Few-shot Random & 91.60 & 91.80 & 93.80 & 91.00 & 91.60 & 93.00 & 89.40 & 91.74   \\
        & & Few-shot TopK & 93.20 & 92.00 & 93.60 & 90.40 & 91.00 & 92.60 & 92.00 & 92.11    \\
        & & Few-shot DPP  & 93.60 & 91.80 & 92.40 & 91.20 & 90.20 & \textbf{93.80} & 91.20 & 92.17    \\
        \cmidrule{3-11}
        & & \cellcolor{gray!20}\OURS         & \cellcolor{gray!20}\textbf{94.60} & \cellcolor{gray!20}\textbf{93.20} & \cellcolor{gray!20}\textbf{94.60} & \cellcolor{gray!20}\textbf{93.80} & \cellcolor{gray!20}\textbf{92.60} & \cellcolor{gray!20}93.60 & \cellcolor{gray!20}\textbf{94.20} & \cellcolor{gray!20}\textbf{93.80}    \\
        \cmidrule{2-11}
        & \multirow{4}{*}{\begin{tabular}[c]{@{}c@{}}\textbf{MedMCQA}\\ (Zs: 52.00) \end{tabular}} 
        & Few-shot Random & 54.80 & 53.60 & 55.80 & 55.40 & 54.00 & 55.00 & 54.00 & 54.66     \\
        & & Few-shot TopK & 54.80 & 55.20 & 55.40 & 55.00 & 54.00 & 54.80 & 54.60 & 54.83   \\
        & & Few-shot DPP  & 56.60 & 53.40 & 54.20 & 55.00 & 54.80 & \textbf{55.40} & 54.80 & 54.89  \\
        \cmidrule{3-11}
        & & \cellcolor{gray!20}\OURS         & \cellcolor{gray!20}\textbf{60.00} & \cellcolor{gray!20}\textbf{56.00} & \cellcolor{gray!20}\textbf{57.60} & \cellcolor{gray!20}\textbf{58.00} & \cellcolor{gray!20}\textbf{57.00} & \cellcolor{gray!20}\textbf{55.40} & \cellcolor{gray!20}\textbf{55.00} & \cellcolor{gray!20}\textbf{57.00}    \\
        \cmidrule{2-11}
        & \multirow{4}{*}{\begin{tabular}[c]{@{}c@{}}\textbf{SciQ}\\ (Zs: 93.60) \end{tabular}} 
        & Few-shot Random & 94.20 & 93.60 & 93.60 & 93.00 & 93.40 & 94.20 & 93.40 & 93.63  \\
        & & Few-shot TopK & 94.20 & 93.20 & 94.20 & 93.20 & 93.80 & 93.80 & 93.80 & 93.74  \\
        & & Few-shot DPP  & 94.40 & 93.00 & 94.00 & 94.00 & 93.60 & 93.80 & 93.20 & 93.71 \\
        \cmidrule{3-11}
        & & \cellcolor{gray!20}\OURS         & \cellcolor{gray!20}\textbf{96.40} & \cellcolor{gray!20}\textbf{94.20} & \cellcolor{gray!20}\textbf{95.80} & \cellcolor{gray!20}\textbf{94.40} & \cellcolor{gray!20}\textbf{95.80} & \cellcolor{gray!20}\textbf{95.00} & \cellcolor{gray!20}\textbf{95.00} & \cellcolor{gray!20}\textbf{95.23}   \\
        \cmidrule{2-11}
        & \multirow{4}{*}{\begin{tabular}[c]{@{}c@{}}\textbf{Social-i-QA}\\ (Zs: 76.00) \end{tabular}} 
        & Few-shot Random & 76.60 & 75.60 & 75.40 & 77.00 & 77.00 & 76.00 & 75.40 & 76.14    \\
        & & Few-shot TopK & 77.00 & 76.20 & 76.60 & 77.20 & 77.00 & 76.00 & 75.80 & 76.54   \\
        & & Few-shot DPP  & 77.00 & 77.40 & 75.80 & 76.60 & 76.20 & 75.60 & 76.20 & 76.40   \\
        \cmidrule{3-11}
        & & \cellcolor{gray!20}\OURS         & \cellcolor{gray!20}\textbf{79.20} & \cellcolor{gray!20}\textbf{78.80} & \cellcolor{gray!20}\textbf{77.60} & \cellcolor{gray!20}\textbf{79.40} & \cellcolor{gray!20}\textbf{77.80} & \cellcolor{gray!20}\textbf{76.80} & \cellcolor{gray!20}\textbf{77.60} & \cellcolor{gray!20}\textbf{78.17}   \\
        \midrule
        \multirow{25}{*}{\begin{tabular}[c]{@{}c@{}}\rotatebox{90}{\textbf{Qwen2.5-7B-Instruct}}\end{tabular}} & \multirow{4}{*}{\begin{tabular}[c]{@{}c@{}}\textbf{ARC-Challenge}\\ (Zs: 84.60) \end{tabular}} 
        & Few-shot Random & 89.40 & 89.90 & 89.60 & 90.00 & 89.60 & 89.40 & 89.00 & 89.56  \\
        & & Few-shot TopK & 90.20 & 90.00 & 89.80 & 89.80 & 89.40 & 88.60 & 89.60 & 89.63 \\
        & & Few-shot DPP  & 89.80 & 89.40 & 89.80 & 90.00 & 89.80 & 89.40 & 89.80 & 89.71   \\
        \cmidrule{3-11}
        & & \cellcolor{gray!20}\OURS         & \cellcolor{gray!20}\textbf{92.60} & \cellcolor{gray!20}\textbf{92.00} & \cellcolor{gray!20}\textbf{92.20} & \cellcolor{gray!20}\textbf{91.40} & \cellcolor{gray!20}\textbf{90.60} & \cellcolor{gray!20}\textbf{90.20} & \cellcolor{gray!20}\textbf{90.60} & \cellcolor{gray!20}\textbf{91.37}  \\
        \cmidrule{2-11}
        & \multirow{4}{*}{\begin{tabular}[c]{@{}c@{}}\textbf{Financial Phrasebank}\\ (Zs: 89.20) \end{tabular}} 
        & Few-shot Random & 91.80 & 92.80 & 92.60 & \textbf{92.60} & 88.80 & 93.40 & 89.60 & 91.66  \\
        & & Few-shot TopK & 92.80 & 91.40 & 92.40 & 92.20 & 89.40 & 93.80 & 91.40 & 91.91  \\
        & & Few-shot DPP  & 92.00 & 92.40 & 92.00 & \textbf{92.60} & 87.40 & 94.00 & 90.20 & 91.51 \\
        \cmidrule{3-11}
        & & \cellcolor{gray!20}\OURS         & \cellcolor{gray!20}\textbf{93.00} & \cellcolor{gray!20}\textbf{93.80} & \cellcolor{gray!20}\textbf{94.00} & \cellcolor{gray!20}\textbf{92.60} & \cellcolor{gray!20}\textbf{92.60} & \cellcolor{gray!20}\textbf{94.60} & \cellcolor{gray!20}\textbf{93.80} & \cellcolor{gray!20}\textbf{93.49}   \\
        \cmidrule{2-11}
        & \multirow{4}{*}{\begin{tabular}[c]{@{}c@{}}\textbf{MedMCQA}\\ (Zs: 53.00) \end{tabular}} 
        & Few-shot Random & 56.60 & 54.80 & 56.40 & 54.60 & 54.00 & 54.40 & 55.20 & 55.14  \\
        & & Few-shot TopK & 54.80 & 54.00 & 57.40 & 55.80 & 54.20 & 54.20 & 55.00 & 55.06  \\
        & & Few-shot DPP  & 55.60 & 55.20 & 56.40 & 54.20 & 54.40 & 55.00 & 55.60 & 55.20  \\
        \cmidrule{3-11}
        & & \cellcolor{gray!20}\OURS         & \cellcolor{gray!20}\textbf{57.60} & \cellcolor{gray!20}\textbf{58.00} & \cellcolor{gray!20}\textbf{58.60} & \cellcolor{gray!20}\textbf{56.20} & \cellcolor{gray!20}\textbf{56.80} & \cellcolor{gray!20}\textbf{56.20} & \cellcolor{gray!20}\textbf{56.20} & \cellcolor{gray!20}\textbf{57.09}  \\
        \cmidrule{2-11}
        & \multirow{4}{*}{\begin{tabular}[c]{@{}c@{}}\textbf{SciQ}\\ (Zs: 92.40) \end{tabular}} 
        & Few-shot Random & 94.40 & 92.80 & 93.40 & 93.60 & 93.80 & 93.80 & 93.80 & 93.66    \\
        & & Few-shot TopK & 94.40 & 93.40 & 93.60 & 94.00 & 93.40 & 93.60 & 93.40 & 93.69   \\
        & & Few-shot DPP  & 94.20 & 93.60 & 93.40 & 94.40 & 93.80 & 94.00 & 93.60 & 93.86   \\
        \cmidrule{3-11}
        & & \cellcolor{gray!20}\OURS         & \cellcolor{gray!20}\textbf{95.00} & \cellcolor{gray!20}\textbf{94.40} & \cellcolor{gray!20}\textbf{95.00} & \cellcolor{gray!20}\textbf{95.80} & \cellcolor{gray!20}\textbf{95.20} & \cellcolor{gray!20}\textbf{96.20} & \cellcolor{gray!20}\textbf{95.00} & \cellcolor{gray!20}\textbf{95.23} \\
        \cmidrule{2-11}
        & \multirow{4}{*}{\begin{tabular}[c]{@{}c@{}}\textbf{Social-i-QA}\\ (Zs: 75.00) \end{tabular}} 
        & Few-shot Random & 76.20 & 75.00 & 75.20 & 77.20 & 77.20 & 75.20 & 75.20 & 75.89   \\
        & & Few-shot TopK & \textbf{77.60} & 75.80 & 75.40 & 77.60 & 75.40 & 76.00 & 76.80 & 76.37   \\
        & & Few-shot DPP  & \textbf{77.60} & 76.60 & 76.40 & 77.80 & 75.60 & 75.00 & 76.00 & 76.43  \\
        \cmidrule{3-11}
        & & \cellcolor{gray!20}\OURS         & \cellcolor{gray!20}\textbf{77.60} & \cellcolor{gray!20}\textbf{77.40} & \cellcolor{gray!20}\textbf{76.60} & \cellcolor{gray!20}\textbf{78.00} & \cellcolor{gray!20}\textbf{77.40} & \cellcolor{gray!20}\textbf{78.40} & \cellcolor{gray!20}\textbf{78.20} & \cellcolor{gray!20}\textbf{77.66}  \\
        \bottomrule
    \end{tabular}
    }
    \label{tab:main-qwen-cd}
\end{table*}

%% file: tables/generation.tex
\begin{table*}[t]
    \centering
    \caption{Experiments on Generation Tasks.}
    \resizebox{0.5\textwidth}{!}{
        \begin{tabular}{l|cc}
        \toprule
        Methods & LiveCodebench & GPQA \\
        \midrule
        Zero-shot       & 16.67 & 35.29 \\
        Few-shot Random & 11.74 & 31.14 \\
        Few-shot TopK   & 13.31 & 31.68 \\
        Few-shot DPP    & 13.89 & 32.97 \\
        \midrule
        \OURS           & \textbf{20.55} & \textbf{40.48} \\
        \bottomrule
        \end{tabular}
    }
    \label{tab:generation_task}
\end{table*}

%% file: tables/injection_position.tex
\begin{table*}[t]
    \centering
    \caption{Performance comparison across different injection positions.}
    \resizebox{0.6\textwidth}{!}{
        \begin{tabular}{l|cccc}
        \toprule
        Position & Random & All & First & Last \\
        \midrule
        \rowcolor{gray!20}\multicolumn{5}{c}{Source Task: Com-QA, Target Task: ARC-C} \\
        \midrule
        Llama3.1-8B          & 75.20 & 75.00 & 74.00 & \textbf{77.20} \\
        Llama3.1-8B-Instruct & 79.60 & 78.20 & 78.80 & \textbf{82.20} \\
        \midrule
        \rowcolor{gray!20}\multicolumn{5}{c}{Source Task: ARC-E, Target Task: MedMCQA} \\
        \midrule
        Llama3.1-8B          & 53.40 & 52.00 & 51.80 & \textbf{57.20} \\
        Llama3.1-8B-Instruct & 53.20 & 57.20 & 57.60 & \textbf{62.00} \\
        \bottomrule
        \end{tabular}
    }
    \label{tab:injection_position}
\end{table*}

%% file: sections/appendix/datasets.tex
\section{Dataset Details}
\label{app:dataset}

In this part, we provide detailed descriptions of the datasets used in our experiments, covering both cross-domain and cross-lingual scenarios.

\subsection{Cross-domain Scenarios}

$\bullet$ \textbf{\underline{ARC-Easy}}: 
ARC-Easy~\cite{ARC} is a multiple-choice question-answering dataset, which consists of simple science exam questions from grade 3 to grade 9. 
These questions are designed to be straightforward and require basic knowledge.

$\bullet$ \textbf{\underline{AG-news}}:
AG-news~\cite{AG-news} is a news topic classification dataset, which is constructed by collecting article titles and descriptions from the four main categories: World, Sports, Business, and Sci/Tech.

$\bullet$ \textbf{\underline{BoolQ}}:
BoolQ~\cite{BoolQ} is a reading comprehension dataset with yes/no questions.
The task requires answering these binary questions based on the given passages.

$\bullet$ \textbf{\underline{Commonsense-QA}}:
Commonsense-QA~\cite{Commonsense-QA} is a multiple-choice question answering dataset that requires different types of commonsense knowledge to find the correct answers.

$\bullet$ \textbf{\underline{MNLI}}:The Multi-Genre Natural Language Inference~(MNLI)~\cite{MNLI} is a crowd-sourced collection of 433k sentence pairs annotated with textual entailment information.
The task is to classify the relationship between two sentences as entailment, contradiction, or neutral.

$\bullet$ \textbf{\underline{QQP}}:
Quora Question Pairs~(QQP)~\cite{QQP} is a natural language understanding dataset comprising over 400k question pairs.
Each question pair is annotated with a binary label indicating whether the two questions are duplicates of each other.

$\bullet$ \textbf{\underline{SST2}}:
The Stanford Sentiment Treebank~(SST2)~\cite{SST2} is a binary sentiment classification dataset, which contains the movie reviews labeled as either positive or negative.

$\bullet$ \textbf{\underline{ARC-Challenge}}:
ARC-Challenge~\cite{ARC} is a more difficult version of ARC-Easy. 
It also includes science exam questions for grades 3 to 9, but requires deeper reasoning and advanced problem-solving strategies.

$\bullet$ \textbf{\underline{Financial Phrasebank}}:
Financial Phrasebank~\cite{Financial-Phrasebank} is a sentiment analysis dataset focused on financial news, which consists of financial news articles annotated with sentiment labels such as positive, negative, or neutral.

$\bullet$ \textbf{\underline{MedMCQA}}:
MedMCQA~\cite{MedMCQA} is a large-scale, multiple-choice question answering dataset, designed to address real-world medical entrance exam questions.

$\bullet$ \textbf{\underline{SciQ}}:
SciQ~\cite{SciQ} is a multiple-choice question answering dataset comprising science exam questions in the fields of physics, chemistry, and biology.

$\bullet$ \textbf{\underline{Social-i-QA}}:
Social-i-QA~\cite{Social-i-QA} is a question-answering benchmark designed to evaluate social commonsense intelligence, which focuses on understanding people’s actions and their social implications.

\subsection{Cross-lingual Scenarios}

$\bullet$ \textbf{\underline{MARC}}:The Multilingual Amazon Reviews Corpus~(MARC)~\cite{MARC} is a large-scale collection of Amazon reviews for multilingual text classification, which contains reviews in six languages: English, Japanese, German, French, Spanish, and Chinese.

%% file: sections/appendix/limitation.tex
\section{Limitation}
\label{app:limitation}

In this paper, we analyze the zero-shot and few-shot prompting across domains from the perspective of activations and propose a new cross-task transfer method \OURS, which utilizes the activations from the high-resource tasks to address low-resource tasks within the latent space of LLMs.
One limitation of our work is that our method requires access to the internal representations of the model, making it infeasible for closed-source LLMs.
In addition, due to resource constraints, we conduct experiments on several representative LLMs, such as Llama and Qwen.
In future work, we hope to explore the effectiveness of our method on other model families as well as vision–language models.

%% file: sections/appendix/case_study.tex
\newpage
\section{Case Study}

\input{tables/case1}

\input{tables/case2}
\input{tables/case3}
\input{tables/case4}

%% file: tables/case1.tex
\begin{center}
\begin{tcolorbox}[colback=blue!5!white,colframe=blue!55!black,width=0.98\linewidth,title={\textit{Case Study 1}}]
{
    {
        \textbf{Zero-shot:} \\
        Definition: Given a question answering task from the 3rd to 9th-grade science exam. The question contains four options "A.", "B.", "C." and "D." Select the most appropriate choice that answers the question. \\
        Question: A student mixed 25 grams of salt into 1,000 grams of water. What is the mass of the saltwater mixture? \\
        A. 975 grams \\ 
        B. 1,000 grams \\
        C. 1,025 grams \\
        D. 2,500 grams \\
        Answer: \textcolor{red}{B} \\ \\
        
        \textbf{Few-shot DPP:} \\
        Definition: Given a context and a question do binary true and false type text classification. You are given a passage as context and a question related to the passage that can be answered as "True" or "False". Based on the context, question and your reasoning ability answer in a "True" and "False". \\
        Context: Usually, the relationship between mass and weight on Earth is highly proportional; objects that are a hundred times more massive than a one-liter bottle of soda almost always weigh a hundred times more--approximately 1,000 newtons, which is the weight one would expect on Earth from an object with a mass slightly greater than 100 kilograms. Yet, this is not always the case and there are familiar objects that violate this mass / weight proportionality. \\
        Question: Is mass the same as weight on earth?\\
        Label:False \\
        ... \\
        
        Definition: Given a question answering task from the 3rd to 9th-grade science exam. The question contains four options "A.", "B.", "C." and "D." Select the most appropriate choice that answers the question. \\
        Question: A student mixed 25 grams of salt into 1,000 grams of water. What is the mass of the saltwater mixture? \\
        A. 975 grams \\
        B. 1,000 grams \\
        C. 1,025 grams \\
        D. 2,500 grams \\
        Answer: \textcolor{red}{A} \\ \\

        \textbf{Ours:} \\
        Definition: Given a question answering task from the 3rd to 9th-grade science exam. The question contains four options "A.", "B.", "C." and "D." Select the most appropriate choice that answers the question. \\
        Question: A student mixed 25 grams of salt into 1,000 grams of water. What is the mass of the saltwater mixture? \\
        A. 975 grams \\
        B. 1,000 grams \\
        C. 1,025 grams \\
        D. 2,500 grams \\
        Answer: \textcolor{green}{C}
    }
}
\end{tcolorbox}
\end{center}

%% file: tables/case2.tex
\begin{center}
\begin{tcolorbox}[colback=blue!5!white,colframe=blue!55!black,width=0.98\linewidth,title={\textit{Case Study 2}}]
{
    {
        \textbf{Zero-shot:} \\
        Definition: Given a sentence mined from a financial news article, you are to determine the sentiment polarity of the sentence. The task deals with financial sentiment analysis. Based on the sentiment conveyed by the sentence, label the sentence as "negative", "positive" or "neutral". \\
        Sentence: Equipment will be manufactured in Vaahto 's workshop in Hollola , Finland and is scheduled for shipments during the first quarter of 2009 . \\
        Label: \textcolor{red}{positive} \\ \\
        
        \textbf{Few-shot DPP:} \\
        Definition: Given a context and a question do binary true and false type text classification. You are given a passage as context and a question related to the passage that can be answered as "True" or "False". Based on the context, question and your reasoning ability answer in a "True" and "False". \\
        Context: Harley-Davidson India is a wholly owned subsidiary of Harley-Davidson, based in Gurgaon, Haryana, India. Harley-Davidson India commenced operations in August 2009 and appointed its first dealership in July 2010.
        Question: does harley davidson have a plant in india
        Label:True \\
        ... \\
        
        Definition: Given a sentence mined from a financial news article, you are to determine the sentiment polarity of the sentence. The task deals with financial sentiment analysis. Based on the sentiment conveyed by the sentence, label the sentence as "negative", "positive" or "neutral". \\
        Sentence: Equipment will be manufactured in Vaahto 's workshop in Hollola , Finland and is scheduled for shipments during the first quarter of 2009 . \\
        Label: \textcolor{red}{positive} \\ \\
        
        \textbf{Ours:} \\
        Definition: Given a sentence mined from a financial news article, you are to determine the sentiment polarity of the sentence. The task deals with financial sentiment analysis. Based on the sentiment conveyed by the sentence, label the sentence as "negative", "positive" or "neutral". \\
        Sentence: Equipment will be manufactured in Vaahto 's workshop in Hollola , Finland and is scheduled for shipments during the first quarter of 2009 . \\
        Label: \textcolor{green}{neutral}
    }
}
\end{tcolorbox}
\end{center}

%% file: tables/case3.tex
\begin{center}
\begin{tcolorbox}[colback=blue!5!white,colframe=blue!55!black,width=0.98\linewidth,title={\textit{Case Study 3}}]
{
    {
        \textbf{Zero-shot:} \\
        Definition: Given a multiple choice question containing four options "A.", "B.", "C." and "D." from a medical entrance exam. The question is related to a sub-field of medical science like Microbiology, Radiology, Ophthalmology, Surgery, Human anatomy, etc. Based on the question, the option and your knowledge of the medical field select the most appropriate answer from the provided choices "A.", "B.", "C." and "D.". \\
        Question: Which of the following is not a component of quick SOFA (qSOFA) scoring?  \\
        A. Bilateral undilated pupils \\
        B. Altered Mentation \\
        C. Glasgow Coma Score \\
        D. SBP <= 100 mm Hg \\
        Answer: \textcolor{red}{C} \\ \\
        
        \textbf{Few-shot DPP:} \\
        Definition: Given a context and a question do binary true and false type text classification. You are given a passage as context and a question related to the passage that can be answered as "True" or "False". Based on the context, question and your reasoning ability answer in a "True" and "False". \\
        Context: The series debuted on January 26, 2017 to positive reviews. A 22-episode second season premiered on October 11, 2017, and concluded on May 16, 2018. On April 2, 2018, The CW renewed the series for a third season, which is set to premiere October 10, 2018. \\
        Question: is there going to be any more episodes of riverdale \\
        Label:True \\
        ... \\
        
        Definition: Given a multiple choice question containing four options "A.", "B.", "C." and "D." from a medical entrance exam. The question is related to a sub-field of medical science like Microbiology, Radiology, Ophthalmology, Surgery, Human anatomy, etc. Based on the question, the option and your knowledge of the medical field select the most appropriate answer from the provided choices "A.", "B.", "C." and "D.". \\
        Question: Which of the following is not a component of quick SOFA (qSOFA) scoring? \\
        A. Bilateral undilated pupils \\
        B. Altered Mentation \\
        C. Glasgow Coma Score \\
        D. SBP <= 100 mm Hg \\
        Answer: \textcolor{red}{C} \\ \\

        \textbf{Ours:} \\
        Definition: Given a multiple choice question containing four options "A.", "B.", "C." and "D." from a medical entrance exam. The question is related to a sub-field of medical science like Microbiology, Radiology, Ophthalmology, Surgery, Human anatomy, etc. Based on the question, the option and your knowledge of the medical field select the most appropriate answer from the provided choices "A.", "B.", "C." and "D.". \\
        Question: Which of the following is not a component of quick SOFA (qSOFA) scoring? \\
        A. Bilateral undilated pupils \\
        B. Altered Mentation \\
        C. Glasgow Coma Score \\
        D. SBP <= 100 mm Hg \\
        Answer: \textcolor{green}{A}
    }
}
\end{tcolorbox}
\end{center}

%% file: tables/case4.tex
\begin{center}
\begin{tcolorbox}[colback=blue!5!white,colframe=blue!55!black,width=0.98\linewidth,title={\textit{Case Study}}]
{
    {
        \textbf{Zero-shot:} \\
        Definition: Given a question from a scientific exam about Physics, Chemistry, and Biology, among others. The question is in multiple choice format with four answer options "A.", "B.", "C." and "D.". Using your knowledge about the scientific fields answer the question and provide the label "A", "B", "C" and "D" as answer. \\
        Question: What happens to energy when work is done by a system? \\
        A. removed \\
        B. stored \\
        C. multiplied \\
        D. added \\
        Answer: \textcolor{red}{B} \\ \\
        
        \textbf{Few-shot DPP:} \\
        Definition: Given a context and a question do binary true and false type text classification. You are given a passage as context and a question related to the passage that can be answered as "True" or "False". Based on the context, question and your reasoning ability answer in a "True" and "False". \\
        Context: The sixth season of the American ABC fantasy-drama Once Upon a Time was ordered on March 3, 2016. It debuted on September 25, 2016, and concluded on May 14, 2017. In January 2017, it was stated that the sixth season would end the main storyline, and for a seventh season, the series would be softly rebooted with a new storyline. \\
        Question: is there a season six of once upon a time \\
        Label:True \\
        ... \\
        
        Definition: Given a question from a scientific exam about Physics, Chemistry, and Biology, among others. The question is in multiple choice format with four answer options "A.", "B.", "C." and "D.". Using your knowledge about the scientific fields answer the question and provide the label "A", "B", "C" and "D" as answer. \\
        Question: What happens to energy when work is done by a system? \\
        A. removed \\
        B. stored \\
        C. multiplied \\
        D. added \\
        Answer: \textcolor{red}{B} \\ \\
        
        \textbf{Ours:} \\
        Definition: Given a question from a scientific exam about Physics, Chemistry, and Biology, among others. The question is in multiple choice format with four answer options "A.", "B.", "C." and "D.". Using your knowledge about the scientific fields answer the question and provide the label "A", "B", "C" and "D" as answer. \\
        Question: What happens to energy when work is done by a system? \\
        A. removed \\
        B. stored \\
        C. multiplied \\
        D. added \\
        Answer: \textcolor{green}{A}
    }
}
\end{tcolorbox}
\end{center}